\documentclass[10pt,twocolumn,letterpaper]{article}

\usepackage{cvpr}
\usepackage{times}
\usepackage{epsfig}
\usepackage{graphicx}
\usepackage{amsmath}
\usepackage{amssymb}

\usepackage{subcaption}
\usepackage{hhline}
\usepackage{xcolor}
\usepackage{multirow}
\usepackage{tabularx}
\usepackage{array}

\usepackage{authblk}

\newcommand{\sz}{0.102}
\newcommand{\szl}{0.067}
\newcommand{\szm}{-30pt}

\newcommand{\sze}{0.13}
\newcommand{\szle}{0.08}

\usepackage[pagebackref=true,breaklinks=true,letterpaper=true,colorlinks,bookmarks=false]{hyperref}

\cvprfinalcopy 

\def\cvprPaperID{647} 

\ifcvprfinal\fi

\begin{document}

\title{3D Hand Shape and Pose from Images in the Wild}

\author{Adnane Boukhayma$^{1}$, Rodrigo de Bem$^{1,2}$, Philip H.S. Torr$^{1}$\\
$^{1}$ University of Oxford, UK\\
$^{2}$ Federal University of Rio Grande, Brazil\\
{\tt\small \{adnane.boukhayma, rodrigo.andradedebem, philip.torr\}@eng.ox.ac.uk}
}

\maketitle

\begin{abstract}
   We present in this work the first end-to-end deep learning based method that predicts both 3D hand shape and pose from RGB images in the wild. Our network consists of the concatenation of a deep convolutional encoder, and a fixed model-based decoder. Given an input image, and optionally 2D joint detections obtained from an independent CNN, the encoder predicts a set of hand and view parameters. The decoder has two components: A pre-computed articulated mesh deformation hand model that generates a 3D mesh from the hand parameters, and a re-projection module controlled by the view parameters that projects the generated hand into the image domain. We show that using the shape and pose prior knowledge encoded in the hand model within a deep learning framework yields state-of-the-art performance in 3D pose prediction from images on standard benchmarks, and produces geometrically valid and plausible 3D reconstructions.
Additionally, we show that training with weak supervision in the form of 2D joint annotations on datasets of images in the wild, in conjunction with full supervision in the form of 3D joint annotations on limited available datasets allows for good generalization to 3D shape and pose predictions on images in the wild.      
\end{abstract}

\begin{figure*}[t!]
\center
\includegraphics[width=.85\linewidth]{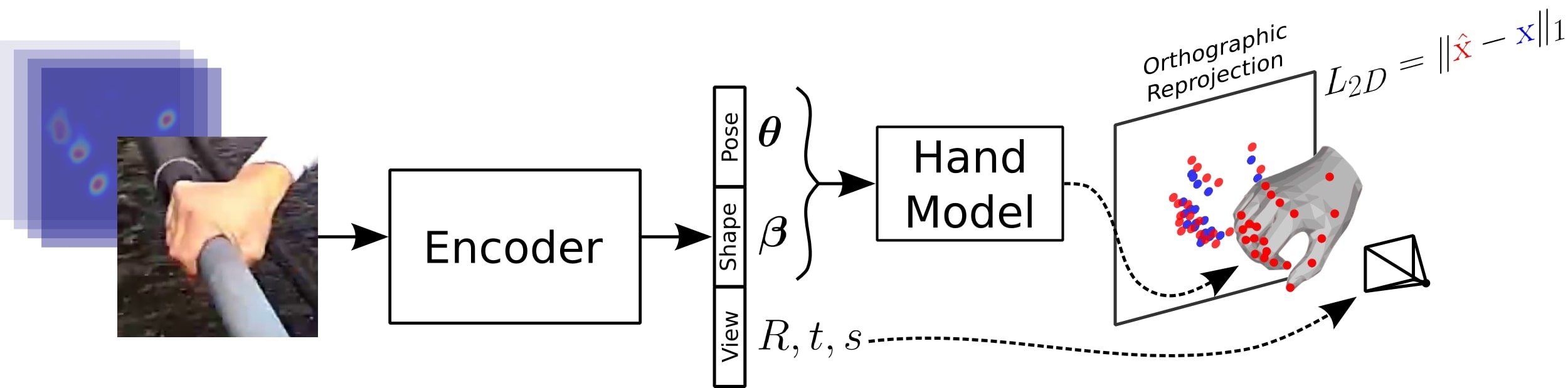}
\vspace{-5pt}
\caption{Our pipeline takes as input a hand image and optionally 2D joint heat-maps from an independent CNN. The encoder generates the shape, pose and view parameters. The hand parameters are fed to the hand model that generates a triangulated 3D mesh and its underlying 3D skeleton. The latter are re-projected into the image domain using a weak perspective camera model controlled by the view parameters. This network is trained end-to-end with a combination of weak 2D and full 3D joint supervision. The hand and view parameters are not supervised.}
\label{fig:pipe}
\end{figure*}

\section{Introduction}
Human hand pose estimation and reconstruction in 3D is a long standing problem in the computer vision and graphics communities that has applications in various domains such as virtual and augmented reality and human-machine interaction \cite{piumsomboon2013user,jang20153d,Song:2014:IGA:2642918.2647373,hurst2013gesture}.
With the abundance of affordable commodity depth cameras, the research literature focused naturally more on estimating 3D hand pose through depth observations (e.g. 
\cite{Wu_2018_ECCV,Zhou_2018_ECCV,Ge_2018_CVPR,Poier2018cvpr_preview,Wan_2018_CVPR}), and many works also explored this problem in multi-view setups 
\cite{panteleris2017back,zhang20163d,rosales20013d,de2006regression,
oikonomidis2010markerless,ellipsoidtracker_3dv2014}. When it comes to a monocular color input, the problem becomes inherently ill posed due to the increased depth and scale ambiguities, but that did not prevent several researchers \cite{1211500,de2011model,stenger2001model,thayananthan2003shape,wu2001capturing,5509753} from attempting to solve it in the past albeit with limited results. More recently, the unprecedented success of deep learning on similar tasks motivated new work with encouraging results for 3D hand pose from single images \cite{zimmermann2017learning,GANeratedHands_CVPR2018,cai2018weakly,spurr2018cross,iqbal2018hand}. Nevertheless, this task remains particularly difficult: Unlike clothed human bodies or faces, hands have an almost uniform appearance and lack characteristic local features such as eyes and mouths in faces. Unlike bodies, they can have more complex pose configurations and they can be captured from a much wider range of views. Furthermore when observed in the wild, as in dataset \textsc{Mpii+Nzsl} \cite{simon2017hand} (Figure \ref{fig:qual}), their images usually contain external occlusion, self-occlusion, clutter and blur due to their motion. Besides, hands are often small in size compared to the scene so cropped patches around them have low resolutions.   
      
The main obstacles for 3D hand pose estimation from images with deep learning include: (i) The lack of large datasets annotated with reliable 3D ground-truth and (ii) the incapability of the current 3D annotated datasets to make networks generalize greatly to challenging images in the wild. 

The first point is tackled by the literature through training with synthetic images \cite{zimmermann2017learning}, populating datasets by transforming synthetic images into real looking ones \cite{GANeratedHands_CVPR2018}, or leveraging auxiliary types of data in training like depth \cite{cai2018weakly,spurr2018cross}. We propose a different and simple yet efficient approach to alleviate both challenges (i) and (ii) by circumventing heavy dependence of 3D data in training: Instead of relying on images paired with 3D joint annotations to learn a prior on hand geometry, we exploit a recently proposed differentiable articulated mesh deformation hand model \cite{romero2017embodied} built with linear blend skinning \cite{Kavan:2005:SBS:1053427.1053429}, and we reformulate the prediction problem into a learning-based model fitting, that can be trained using both 3D and 2D joint annotations. Training with 2D annotated images allows access to larger datasets (e.g. \textsc{Panoptic} \cite{simon2017hand}) with a fair share of annotated images in the wild (e.g. \textsc{Mpii+Nzsl} \cite{simon2017hand}) compared to datasets with 3D ground-truth, thus helping improve generalization to this type of challenging data. Given an input image, and optionally 2D joint detections obtained from an independent CNN, a deep convolutional encoder predicts the hand shape and pose parameters and view parameters. The model-based decoder uses the latter to generate a 3D triangulated hand mesh and its underlying skeleton, along with their re-projection in image domain (see Figure \ref{fig:pipe}). 

Our contributions in this paper are as follows: This work is the first to propose end-to-end learning of both 3D hand shape and pose from a single RGB image. We also show for the first time that the prior knowledge of factored hand shape and pose in a pre-computed linear blend skinning \cite{Kavan:2005:SBS:1053427.1053429} hand model \cite{romero2017embodied} combined with a deep-convolutional encoder yields state-of-the-art performance in 3D pose prediction from images, and produces geometrically valid and plausible 3D reconstructions, without the need for post-processing optimizations \cite{GANeratedHands_CVPR2018}. We show that this strategy combined with training on 2D annotated datasets of images in the wild produces good generalization in 3D hand reconstruction for challenging images in uncontrolled environments.   

We evaluate our work both quantitatively in terms of 3D pose estimation and qualitatively using various public datasets. These evaluation sets account for cases of hand interaction with objects, occlusion and clutter, and contain egocentric view images, third person view images, and images in the wild. Our method obtains state of-the-art results on standard benchmarks, even compared to methods using additional depth information in training \cite{cai2018weakly,spurr2018cross}, camera intrinsics \cite{GANeratedHands_CVPR2018,panteleris2018using}, and post-processing optimization \cite{GANeratedHands_CVPR2018}. Our method shows superior qualitative results on a challenging dataset of images in the wild (Figure \ref{fig:qual} \& supplementary material).

\section{Related work}

There is a rich literature on 3D hand pose and reconstruction from depth 
\cite{Wu_2018_ECCV,Zhou_2018_ECCV,Ge_2018_CVPR,Poier2018cvpr_preview,Wan_2018_CVPR,ge2016robust,
sharp2015accurate,sinha2016deephand,
keskin2012hand,khamis2015learning,li20153d,oberweger2015training,qian2014realtime,
sridhar2015fast,sun2015cascaded,tang2015opening,tompson2014real,xu2013efficient}, 
image and depth \cite{makris2015model,oikonomidis2011efficient,RealtimeHO_ECCV2016,OccludedHands_ICCV2017}, 
stereo
\cite{panteleris2017back,zhang20163d,rosales20013d} 
and multiple images 
\cite{de2006regression,oikonomidis2010markerless,ellipsoidtracker_3dv2014}. We focus hereby on research material that solely considers a single color input image.    
 
\paragraph{3D hand pose from a single image}
\vspace{-10pt}

\subparagraph{Pre-deep learning}
\vspace{-10pt}
There have been attempts to solve 3D hand pose estimation from a monocular color input prior to deep learning with both discriminative and generative approaches  \cite{1211500,de2011model,stenger2001model,thayananthan2003shape,wu2001capturing,5509753}. However, most of these methods have limited performance and depend on various requirements such as careful initialization and prior knowledge of the background.  
  
\subparagraph{Deep learning}
\vspace{-10pt}
The work of \cite{zimmermann2017learning} was the first to propose 3D hand pose estimation from single images using deep learning. Their method consists of the concatenation of three networks that segment the hand, predict 2D joints, and then predict 3D joints subsequently. The work of \cite{GANeratedHands_CVPR2018} shows that the previous method generalizes poorly to real world images since a major part of their training data is synthetic. In turn, they (\cite{GANeratedHands_CVPR2018}) propose to use Cycle-GAN \cite{zhu2017unpaired} to transform synthetic 3D annotated images of hands into real looking ones. The resulting data is used to train a regressor to predict 2D and 3D hand joints. A final optimization step fits a 3D skeleton to the former 2D and 3D predictions using the camera intrinsics. 
The method in \cite{panteleris2018using} consists in an optimization that fits a hand model to 2D joint detections obtained from a state-of-the-art CNN \cite{simon2017hand}. We also use a pre-defined hand model \cite{romero2017embodied} but within a pipeline trained end-to-end. 

\subparagraph{Depth regularization}
\vspace{-10pt}
Recent works tackle depth ambiguity in 3D hand pose prediction from images by leveraging depth maps in training. \cite{cai2018weakly} proposes to reduce the dependency on noisy 3D annotations in real datasets by introducing a network that predicts full depth maps from the 3D joints. This depth regularizer is trained with ground-truth depth data for both real and synthetic training images, while the 3D predictions are only supervised by the reliable synthetic labels. The authors in \cite{spurr2018cross} use multiple variational auto-encoders sharing the same latent space each auto-encoding a separate hand data modality (e.g. images, 2D joints, 3D joints). They show that the auxiliary auto-encoders help regularize the latent space and produce improved cross-modal predictions (e.g. image to 3D joints). \cite{iqbal2018hand} shows that predicting an implicit 2.5D heat-map representation yields improved 3D predictions even without explicit full depth-map supervision. 

\paragraph{Hand models}
\vspace{-10pt}
Many hand models have been proposed in the literature primarily aiming at tracking depth and color data, where the hand is modelled using various techniques such as assembled geometric primitives \cite{oikonomidis2011efficient}, sum of Gaussians \cite{ellipsoidtracker_3dv2014}, sphere meshes \cite{tkach2016sphere} or loop subdivision of a control mesh \cite{khamis2015learning}. In order to better capture the shape of the hand,  \cite{oikonomidis2011efficient} defines scaling terms to allow bone length to vary, while \cite{taylor2014user} pre-calibrates the shape to fit the hand of interest. The work in \cite{khamis2015learning} was the first to learn hand shape variation from scans with linear blend skinning \cite{Kavan:2005:SBS:1053427.1053429}. The model proposed recently in \cite{romero2017embodied} and referred to as MANO improves on the latter by learning pose dependent corrective blend shapes \cite{SMPL:2015}, thus modelling both hand shape and pose and generating more realistic posed meshes. We use the MANO \cite{romero2017embodied} model in this work.    

\paragraph{Model-based decoders}
\vspace{-10pt}
Several works propose to combine deep convolutional encoders with generative models as decoders for human face \cite{tewari2017mofa,tewari2017self} and body \cite{kanazawa2018end,tung2017self} 3D reconstruction. In many of these works, the decoder is a combination of a parametric model (e.g. linear face model \cite{blanz1999morphable}, SMPL \cite{SMPL:2015}) and a re-projection/rendering module. While most works fix these decoders, some propose to tune them after a supervised initialization \cite{abrevaya2018multilinear,laine2017production,tewari2017self}. This is the first work to propose a combination of a CNN encoder with a fixed generative hand model \cite{romero2017embodied} for the problem of 3D hand reconstruction from images.

\section{Overview}

As illustrated in Figure \ref{fig:pipe}, our pipeline takes as input an image of a hand and optionally 2D joint heat-maps from an independent hand detector. A deep convolutional encoder processes the input and generates a set of hand shape $\boldsymbol{\beta}$ and pose $\boldsymbol{\theta}$ parameters, and a set of view parameters $R$, $t$ and $s$. The hand parameters are fed to a differentiable articulated mesh deformation hand model that generates a triangulated 3D mesh and its underlying 3D skeleton. These outputs are then re-projected into the image domain through a weak perspective camera model controlled by the view parameters. The re-projection module and the hand model together form a model-based decoder whose parameters are fixed and do not require training. The encoder is pre-trained with synthetic examples that we created as elaborated in Section \ref{sec:encoder}. We note that the training of our pipeline is done end-to-end using 2D and 3D joint annotations without supervision on the hand and view parameters, except for a regularization on the hand parameters to ensure their magnitude is small. We detail and explain the functioning of the various parts of the pipeline in the following.  

\section{Hand model}

We use the MANO hand model \cite{romero2017embodied} which is based on the SMPL model for human bodies \cite{SMPL:2015}. It is an articulated mesh deformation model represented with a differentiable function $M(\boldsymbol{\beta},\boldsymbol{\theta})$ taking as input two sets of parameters $\boldsymbol{\beta}$ and  $\boldsymbol{\theta}$ that control the shape and pose of the generated hand respectively:   
\begin{equation}
M(\boldsymbol{\beta},\boldsymbol{\theta})=W(T(\boldsymbol{\beta},\boldsymbol{\theta}),J(\boldsymbol{\beta}),\boldsymbol{\theta},\mathcal{W}),
\end{equation}
where $W$ is a linear blend skinning \cite{Kavan:2005:SBS:1053427.1053429} function applied to a template hand triangulated mesh $T$ rigged with a kinematic tree of $K=16$ joints. $J$ represents the joint locations and it is learned as a sparse linear regressor from mesh vertices, and $\mathcal{W}$ are the blend weights.

In order to reduce the artifacts of linear blend skinning such as overly smooth outputs and mesh collapse around joints, the hand template $T$ is obtained by deforming a mean mesh $\bar{\mathbf{T}}$ with both shape and pose corrective blend shapes, $\mathbf{S}_n$ and $\mathbf{P}_n$ respectively, as follows: 
\begin{equation}
T(\boldsymbol{\beta},\boldsymbol{\theta})=\bar{\mathbf{T}} +
\sum_{n=1}^{|\boldsymbol{\beta}|}\beta_n\mathbf{S}_n +
\sum_{n=1}^{9K}(R_n(\boldsymbol{\theta})-R_n(\boldsymbol{\theta}^*))\mathbf{P}_n,
\label{eq:def}
\end{equation}
where $R_n(\boldsymbol{\theta})$ is the $n^{th}$ element of a vector concatenating rotation matrix coefficients from all joints for pose $\boldsymbol{\theta}$ and $\boldsymbol{\theta}^*$ is the rest pose. The model constants $\{\bar{\mathbf{T}}, \mathbf{S}, \mathbf{P}, J, \mathcal{W}\}$ are learned using registered hand scans from $31$ subjects performing roughly $51$ hand poses. 

In the SMPL model, the pose vector $\boldsymbol{\theta}$ stacks the angle-axis values of the joints. To help the hand model generate physically plausible poses, the authors in \cite{romero2017embodied} reduce this pose representation to a linear embedding by performing Principal Component Analysis on angle-axis values of the joints in the data collected to build the model. The pose vector $\boldsymbol{\theta}$ contains the resulting main coefficients from PCA instead of the angle-axis values. $10$ coefficients are retained for the pose ($\boldsymbol{\theta}\in{\rm I\!R}^{10}$), and $10$ coefficients are used to represent the shape as well ($\boldsymbol{\beta}\in{\rm I\!R}^{10}$).

Given input shape and pose parameters, we obtain a hand mesh $M(\boldsymbol{\beta},\boldsymbol{\theta})$ of $N=778$ vertices and $1538$ faces, along with the corresponding 3D joints $J(\boldsymbol{\beta},\boldsymbol{\theta}) = R_{\boldsymbol{\theta}}(J(\boldsymbol{\beta}))$ where $R_{\boldsymbol{\theta}}$ is the global rigid transformation induced by pose $\boldsymbol{\theta}$. As the hand skeleton in MANO does not contain finger tip joints, we append $J$ with 5 vertices from the hand mesh that correspond to these key-points. The final 3D joint output $J(\boldsymbol{\beta},\boldsymbol{\theta})$ counts $21$ key-points.


\section{Camera model}

In order to re-project the 3D hand mesh vertices $M(\boldsymbol{\beta},\boldsymbol{\theta})$ and 3D joints $J(\boldsymbol{\beta},\boldsymbol{\theta})$ into the 2D image plane, we use the weak perspective model. This approximation allows us to train with annotated images even in the absence of camera intrinsics, which is the case of images in the wild obtained from Youtube videos for instance (e.g. dataset \textsc{Mpii+Nzsl}). Given a global rotation matrix $R\in SO(3)$, a translation $t\in {\rm I\!R}^2$ and a scaling $s\in {\rm I\!R}^+$, the projection writes:    
\begin{equation}
\hat{\mathrm{x}}=s\Pi(RJ(\boldsymbol{\beta},\boldsymbol{\theta}))+t,
\end{equation}
\begin{equation}
\hat{\mathrm{y}}=s\Pi(RM(\boldsymbol{\beta},\boldsymbol{\theta}))+t,
\end{equation}
where $\Pi$ is the orthographic projection.

\section{Encoder}
\label{sec:encoder}

Given an input hand image, the goal of the encoder is to predict the corresponding hand pose and shape parameters $\{\boldsymbol{\beta},\boldsymbol{\theta}\}$ and camera parameters $\{R,t,s\}$. We use the ResNet-50 network \cite{he2016identity} and we adjust the final fully connected layer to output a vector $v=\{R,t,s,\boldsymbol{\beta},\boldsymbol{\theta}\} \in {\rm I\!R}^{26}$. We note that global rotation $R$ is encoded with axis-angle values and is hence represented with $3$ parameters. We also experiment with feeding 2D hand joint heat-maps obtained with a state of the art method \cite{simon2017hand} as additional channel input to the hand RGB image.
        
\paragraph{Network pre-training} 
\vspace{-10pt}
\begin{figure}[h]
\center
\includegraphics[width=0.21\linewidth]{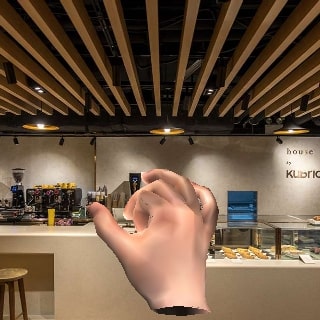}
\includegraphics[width=0.21\linewidth]{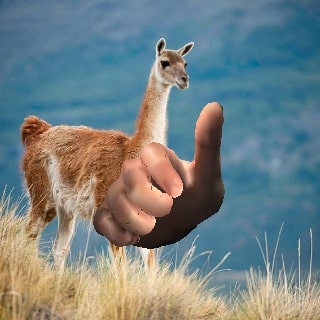}
\includegraphics[width=0.21\linewidth]{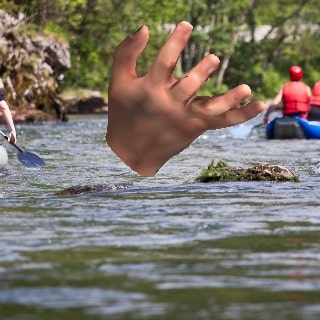}
\includegraphics[width=0.21\linewidth]{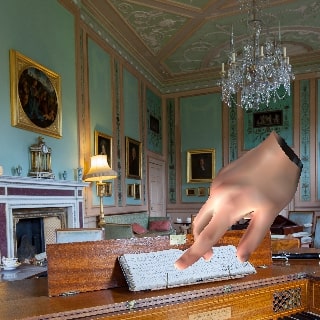}
\par\addvspace{2pt}
\includegraphics[width=0.21\linewidth]{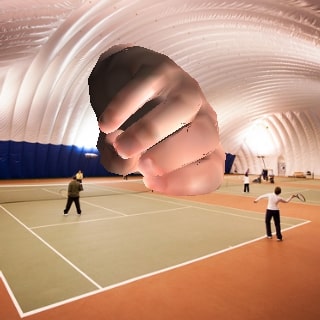}
\includegraphics[width=0.21\linewidth]{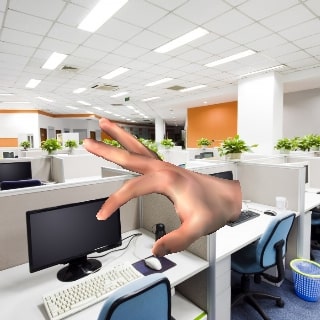}
\includegraphics[width=0.21\linewidth]{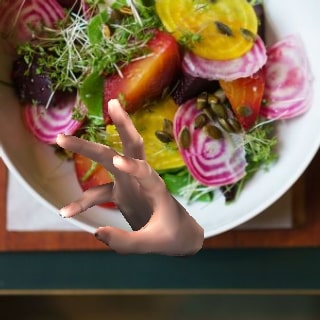}
\includegraphics[width=0.21\linewidth]{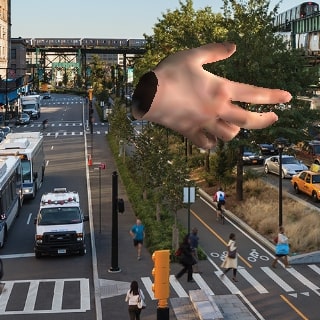}
\caption{Examples from our synthetic dataset created to pre-train the encoder.}
\label{fig:pretrain}
\end{figure}

We pre-train the encoder to ensure that the camera and hand parameters converge towards acceptable values. For this purpose, we create a synthetic dataset of paired hand images with their ground-truth camera and hand parameters using the same generative model that we use as a decoder. Hand geometries are obtained by sampling poses $\boldsymbol{\theta} \in [-2,2]^{10}$ and shapes $\boldsymbol{\beta} \in [-0.03,0.03]^{10}$ then applying rotations $R$, translations $t$ and scalings $s$. Although the work of  \cite{romero2017embodied} does not model hand appearance, the authors provide the scans used to build the geometry model with their registered counterparts. The original scans come with 3D coordinates and RGB values for each vertex. We create example hand appearances using the registered scan topology: To each vertex in a registered mesh, we assign the RGB value of the closest vertex in the original corresponding scan, and we interpolate these values inside faces. The textured hands are finally rendered on top of random background images. Figure \ref{fig:pretrain} shows examples from the resulting dataset.

\section{Training objective}

We combine multiple losses to train our pipeline: A 2D joint re-projection loss $L_{2D}$, a 3D joint loss $L_{3D}$, a hand mask loss $L_{mask}$ and a model parameter regularization loss $L_{reg}$.
\begin{equation}
L=L_{2D} + \alpha_{{\tiny 3D}}L_{3D} + \alpha_{mask}L_{mask} + \alpha_{reg}L_{reg},
\end{equation}
where $\alpha_{3D}=10^2$, $\alpha_{mask}=10^2$ and $\alpha_{reg}=10^1$ are weighting factors. 

\paragraph{2D joint re-projection loss}
\vspace{-10pt} 
This loss ensures that the re-projected hand joints in the image plane coincide with the ground-truth 2D hand joint annotations:  
\begin{equation}
L_{2D} = \|\hat{\mathrm{x}} - \mathrm{x}\|_1,
\end{equation}
where $\mathrm{x}$ is a vector containing the ground-truth 2D hand joint coordinates. We use the $L_1$ loss to account for inaccuracies in hand annotations in our training datasets. 

\paragraph{3D joint loss} 
\vspace{-10pt}
When ground-truth 3D hand joint annotations are available (e.g \textsc{Stereo} dataset), this loss minimises the distance between the latter and the 3D hand joints generated by the hand model:
\begin{equation}
L_{3D} = \|RJ(\boldsymbol{\beta},\boldsymbol{\theta}) - \mathrm{x_{3D}}\|_2^2,
\end{equation}
where $\mathrm{x_{3D}}$ is a vector containing the ground-truth 3D hand joint coordinates.

\paragraph{Hand mask loss}
\vspace{-10pt}

\begin{figure}[h!]
\center
\begin{subfigure}[t]{0.25\linewidth}
\includegraphics[width=\linewidth]{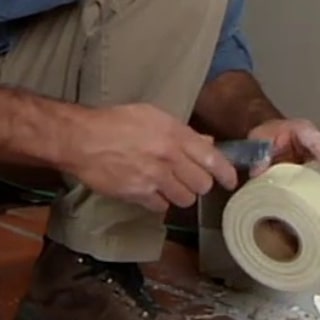}
\end{subfigure}
\begin{subfigure}[t]{0.25\linewidth}
\includegraphics[width=\linewidth]{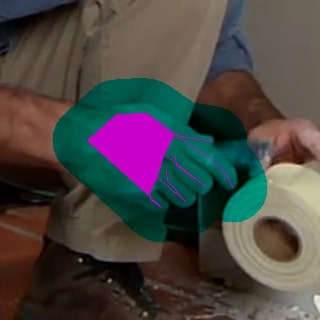}
\end{subfigure}
\begin{subfigure}[t]{0.25\linewidth}
\includegraphics[width=\linewidth]{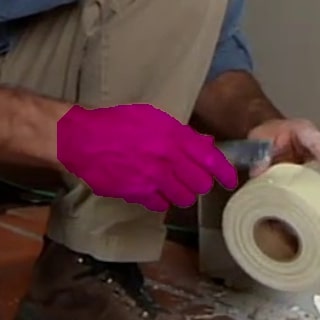}
\end{subfigure}
\par\addvspace{2pt}
\begin{subfigure}[t]{0.25\linewidth}
\includegraphics[width=\linewidth]{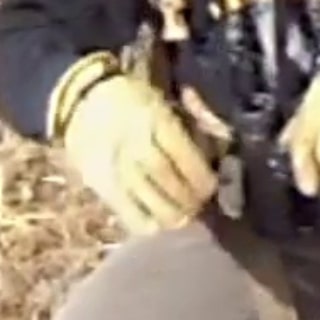}
\caption{}
\label{fig:seg1}
\end{subfigure}
\begin{subfigure}[t]{0.25\linewidth}
\includegraphics[width=\linewidth]{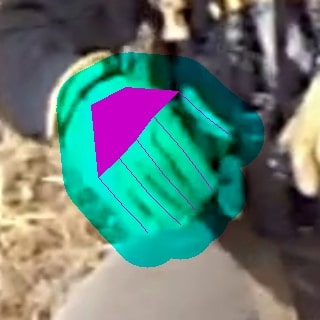}
\caption{}
\label{fig:seg2}
\end{subfigure}
\begin{subfigure}[t]{0.25\linewidth}
\includegraphics[width=\linewidth]{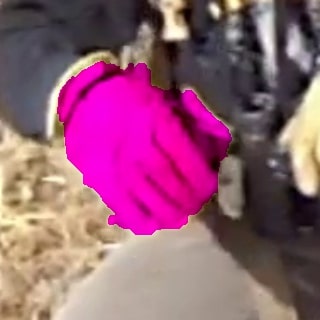}
\caption{}
\label{fig:seg3}
\end{subfigure}
\caption{GrabCut \cite{rother2004grabcut} hand segmentation initialized with 2D joint annotation.
(a) Input image, (b) foreground, background and undecided regions from 2D joints, (c) final segmentation.}
\label{fig:seg}
\end{figure}

We introduce this novel loss to help speed up the convergence of our training and refine hand shape predictions. This loss penalizes re-projected hand vertices that lie outside of the hand region in a binary mask, which is pre-computed prior to training:
\begin{equation}
L_{mask} = 1 - \frac{1}{N} \sum_i H(\hat{\mathrm{y}}_i),
\end{equation}
where $H$ is an occlusion-aware hand mask, i.e $H(u)=1$ if pixel $u$ is inside the hand region even if the hand is occluded in the image, and $H(u)=0$ otherwise. Notice that these masks cannot be obtained with hand skin segmentation methods (e.g.\cite{Li_2013_CVPR,Bambach_2015_ICCV}) as they are sensitive to occlusions.    

We obtain an approximation of these masks (Figure \ref{fig:seg}) for training images using the GrabCut \cite{rother2004grabcut} algorithm, by initializing the foreground, background and probable foreground/background regions using the 2D hand joint annotations: As illustrated in Figure \ref{fig:seg2}, we create an initial foreground by drawing lines of $1$ pixel width connecting joints according to the hand skeleton hierarchy. Pixels inside triangles formed by joints that belong anatomically to the hand surface are appended to the foreground as well. The undecided area is defined as the region within $70$ pixels at most from the foreground, and the remaining pixels are assigned to the initial background.   

\paragraph{Regularization loss}
\vspace{-10pt}
This loss acts on the hand model parameters at the encoder output by reducing their magnitude for physically plausible hand reconstructions and reduced mesh distortions: 
\begin{equation}
L_{reg} = \|\boldsymbol{\theta}\|_2^2 + \alpha_{\boldsymbol{\beta}} \|\boldsymbol{\beta}\|_2^2, 
\label{eq:reg}
\end{equation}
where $\alpha_{\boldsymbol{\beta}}=10^4$ is a weighting factor.

\section{Evaluation}
\label{sec:eval}

We evaluate our method's 3D pose estimates quantitatively and its 3D reconstructions qualitatively on several datasets and with respect to state-of-the-art methods. Without access to camera intrinsics, and trained merely with 2D and 3D joint annotations, our method outperforms deep learning based competing methods, including those using additional depth information in training or camera intrinsics in evaluation. We show particularly superior 3D reconstructions on images in the wild that present challenging situations such as blur, low resolution, occlusion, extremely varying viewpoints and hand pose configurations.

Similar to \cite{simon2017hand}, input images are assumed to be crops of fixed size around the hand. To achieve this, we use a hand key-point detector \cite{simon2017hand} to find the tightest rectangular box of edge size $l$ containing the hand. Images are then cropped with a square patch of size $2.2l$ centred at the same 2D location as the previously detected box. The resulting cropped images are subsequently resized to have a width and height of $320$. As done in \cite{simon2017hand}, we use the right hand model and images of left hands are flipped horizontally.

Finally, we train our pipeline (Figure \ref{fig:pipe}) using the Adam solver \cite{kingma2014adam} with a learning rate of $10^{-4}$ and weight decay of $10^{-5}$.

\paragraph{Datasets}
\vspace{-10pt}
Our training set is made of dataset \textsc{Panoptic} \cite{simon2017hand} that counts $14847$ images, the training set of \textsc{Mpii+Nzsl} \cite{simon2017hand} that counts $1912$ images following the split in \cite{simon2017hand}, and the training set of \textsc{Stereo} \cite{zhang20163d} that counts $15000$ images following the split in \cite{zimmermann2017learning}. This amounts to $31729$ training images, $15000$ (\textsc{Stereo}) with 3D joint annotations, and the remaining $16729$ (\textsc{Panoptic} \& \textsc{Mpii+Nzsl}) with 2D joint annotations only. 

The \textsc{Panoptic} dataset \cite{simon2017hand} contains hands in various poses observed from multiple views in the Panoptic studio \cite{joo2015panoptic}. The \textsc{Mpii+Nzsl} dataset \cite{simon2017hand} is a combination of manually annotated images from The MPII Human Pose dataset \cite{andriluka20142d} containing images from YouTube videos, and images from the New Zealand Sign Language (NZSL) Exercises of the Victoria University of Wellington \cite{NZSL}. The \textsc{Stereo} dataset \cite{zhang20163d} shows an actor's hand in third person view counting with the fingers and moving the hand randomly.

For evaluation, we use the \textsc{Dexter+Object} dataset \cite{RealtimeHO_ECCV2016} which shows interactions of an actor's hand with a cuboid object from a third person view. To evaluate robustness to occlusions and clutter, we use the \textsc{EgoDexter} dataset \cite{OccludedHands_ICCV2017} that displays a hand from an egocentric view interacting with various objects. We finally use the testing set of \textsc{Mpii+Nzsl} \cite{simon2017hand} to asses performance in the presence of blur, low resolution, varying viewpoints and hand pose configurations, among other characteristics of datasets of images in the wild. 

\paragraph{Metrics}
\vspace{-10pt}
To quantitatively evaluate 3D hand pose estimations, we report the percentage of correct points in 3D (3D PCK) and the average 3D Euclidean distance between the estimated 3D joints and the ground-truth when the latter is available, where distances are expressed in millimeters (mm). When only ground-truth 2D joint annotations are available (dataset \textsc{Mpii+Nzsl}), we report 2D PCK and the average 2D Euclidean distance between the estimated 2D re-projected joints and the ground-truth, where distances are expressed in pixels (px).  

\paragraph{Comparison to competing methods}
\vspace{-10pt}

\begin{figure}[h!]
\center
\includegraphics[width=.8\linewidth]{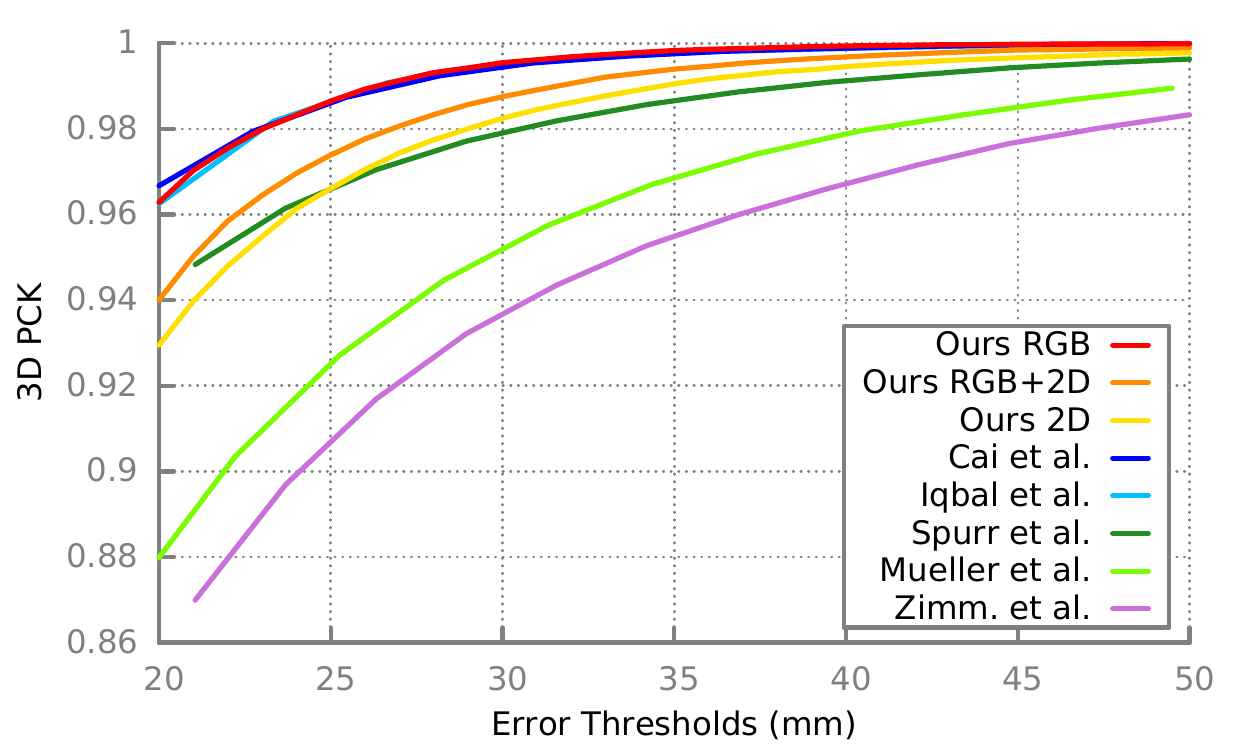}
\vspace{-8pt}
\caption{3D PCK for \textsc{Stereo}.}
\label{fig:pck_s1}
\end{figure}

\begin{figure}[h!]
\center
\includegraphics[width=.8\linewidth]{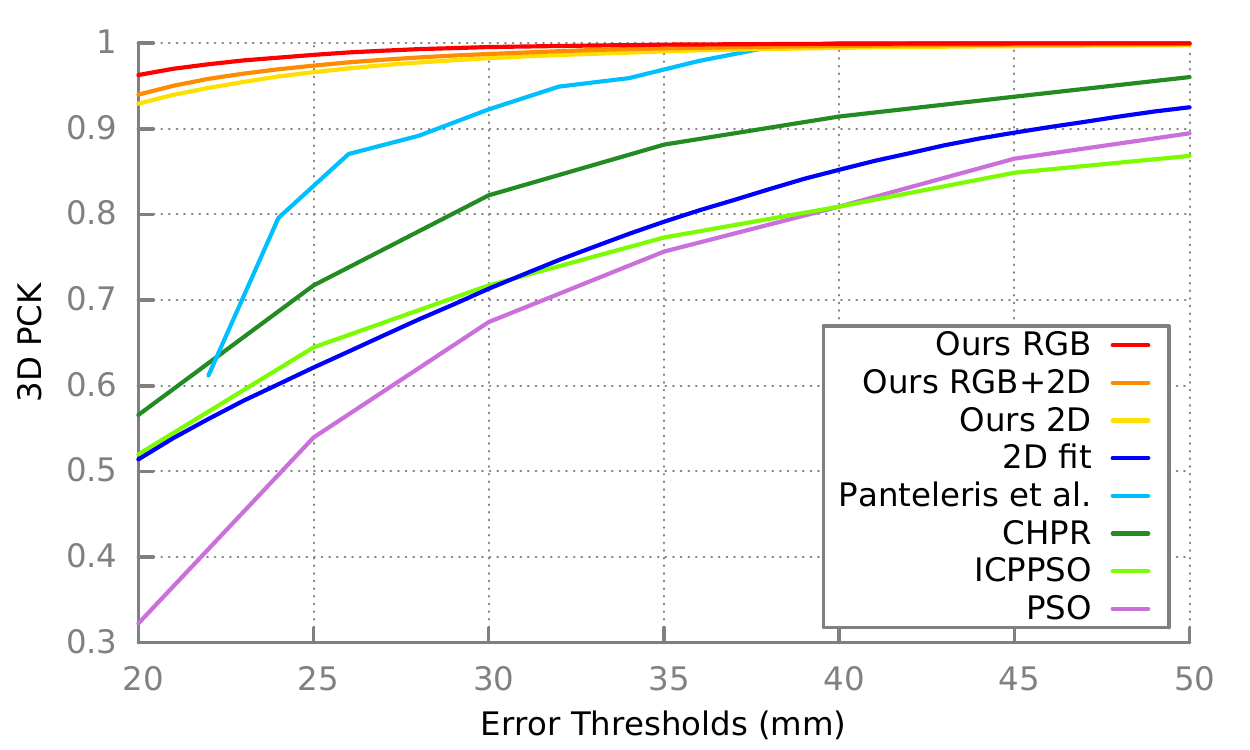}
\vspace{-8pt}
\caption{3D PCK for \textsc{Stereo}.}
\label{fig:pck_s2}
\end{figure}

\vspace{-5pt}
\begin{table}[h!]
\center
\resizebox{0.78\linewidth}{!}{%
\begin{tabular}{|c|*{4}{>{\centering\arraybackslash}m{1.0cm}|}}
\cline{2-5}
\multicolumn{1}{c|}{} & Ours {\footnotesize RGB} & Ours {\footnotesize RGB+2D} & Ours {\footnotesize 2D} & 2D fit\\ 
\hhline{-:=:=:=:=}
3D distance & \textbf{9.76} & 10.18 & 10.46 & 23.21\\
\hline 
\end{tabular}
}
\vspace{-5pt}
\caption{Average 3D joint distance (mm) to ground-truth for \textsc{Stereo}.}
\label{tab:Stereo}
\end{table}

We compare our results on the \textsc{Stereo} dataset to state-of-the-art methods in terms of 3D PCK in Figures \ref{fig:pck_s1} and \ref{fig:pck_s2}, and we show 3D joint errors in Table \ref{tab:Stereo}. Figure \ref{fig:pck_s1} shows deep learning based methods (Cai et al. \cite{cai2018weakly}, Iqbal et al. \cite{iqbal2018hand}, Spurr et al. \cite{spurr2018cross}, Mueller et al. \cite{GANeratedHands_CVPR2018}, Zimm. et al \cite{zimmermann2017learning}) and Figure \ref{fig:pck_s2} shows methods that do not rely on deep learning (Panteleris et al. \cite{panteleris2018using}, PSO, ICPPSO, CHPR \cite{zhang20163d} ). For this experiment, we add a key-point at the center of the hand palm in the MANO model \cite{romero2017embodied} as an interpolation of several mesh vertices to match the annotation of the \textsc{Stereo} dataset. We reproduce the evaluation protocol initially introduced in \cite{zimmermann2017learning} by training on $10$ sequences and testing on the remaining $2$ and aligning predictions to the ground-truth hand root joint. Additionally, for a fair comparison to works \cite{cai2018weakly,spurr2018cross,iqbal2018hand}, we crop the hand images for this experiment such that the final image size is $150\%$ the size of the hand. Using RGB image input only, we obtain state-of-the results even though some of the competing methods use depth data in training (\cite{cai2018weakly,iqbal2018hand}) in addition to images, while others (\cite{GANeratedHands_CVPR2018}) post-process their output with an optimization that fits their hand skeleton to their 3D and 2D joint predictions, and which uses the camera intrinsics as an additional input.

\begin{figure}[h!]
\center
\includegraphics[width=.8\linewidth]{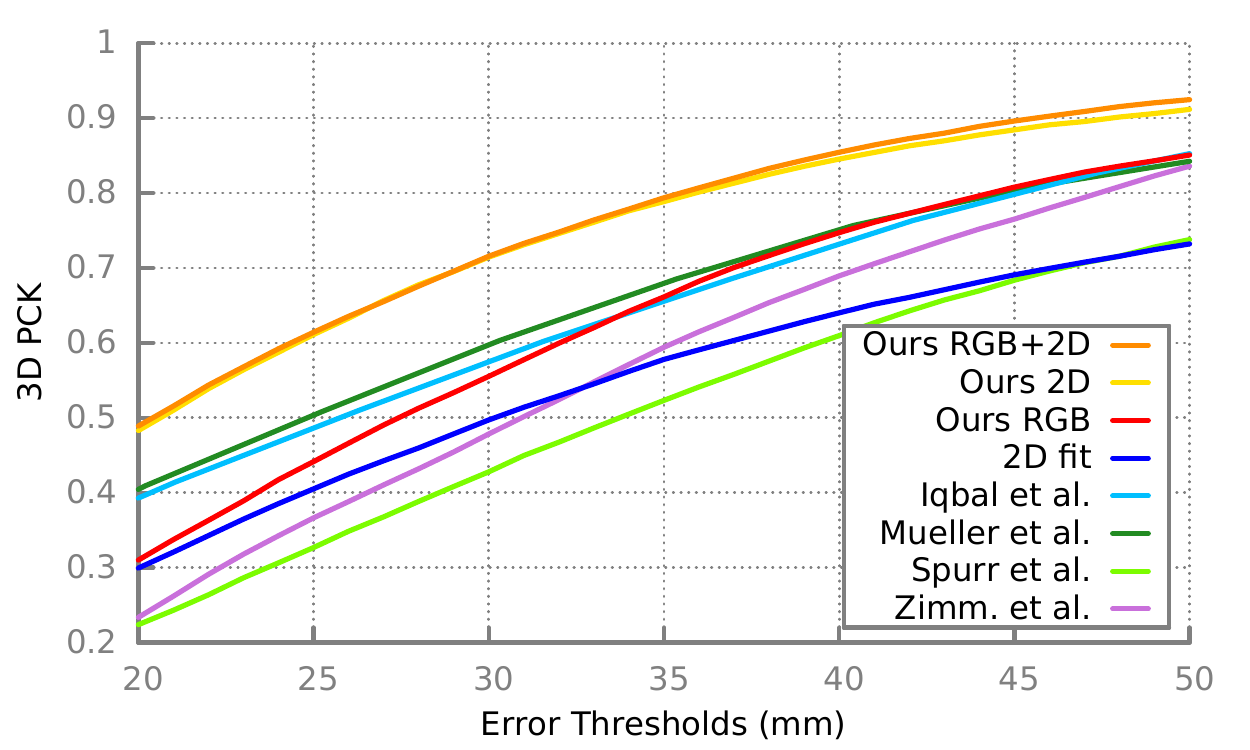}
\vspace{-8pt}
\caption{3D PCK for \textsc{Dexter+Object}.}
\label{fig:pck_DO}
\end{figure}

\vspace{-5pt}
\begin{table}[h!]
\center
\resizebox{\linewidth}{!}{%
\begin{tabular}{|c|*{6}{>{\centering\arraybackslash}m{1.0cm}|}}
\cline{2-7}
\multicolumn{1}{c|}{} & Ours {\footnotesize RGB} & Ours {\footnotesize RGB+2D} & Ours {\footnotesize 2D} & 2D fit & Spurr et al. & Zimm. et al. \\ 
\hhline{-:=:=:=:=:=:=}
3D distance & 33.16 & \textbf{25.53} & 25.93 & 41.18 & 40.20 & 34.75\\
\hline 
\end{tabular}
}
\vspace{-5pt}
\caption{Average 3D joint distance (mm) to ground-truth for \textsc{Dexter+Object}.}
\label{tab:D+O}
\end{table}

\begin{figure}[h!]
\center
\includegraphics[width=.8\linewidth]{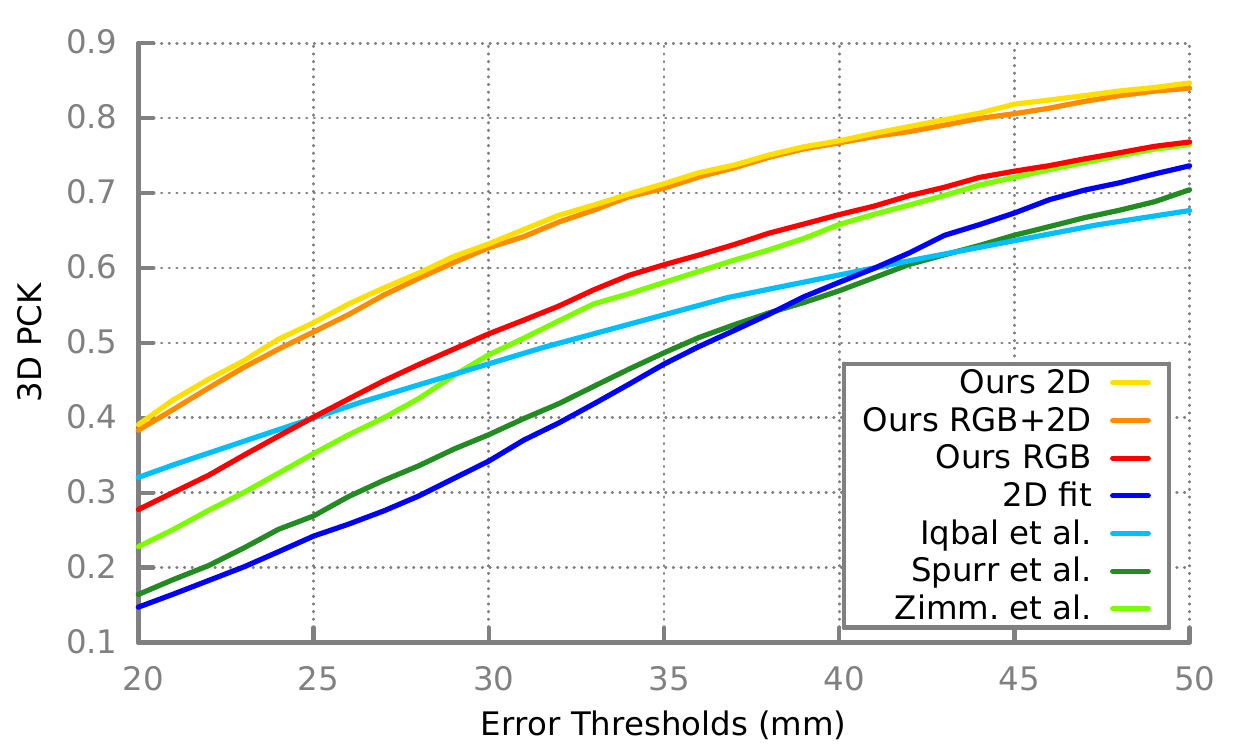}
\vspace{-8pt}
\caption{3D PCK for \textsc{EgoDexter}.}
\label{fig:pck_ED}
\end{figure}
 
\begin{table}[h!]
\center
\resizebox{\linewidth}{!}{%
\begin{tabular}{|c|*{6}{>{\centering\arraybackslash}m{1.0cm}|}}
\cline{2-7}
\multicolumn{1}{c|}{} & Ours {\footnotesize RGB} & Ours {\footnotesize RGB+2D} & Ours {\footnotesize 2D} & 2D fit & Spurr et al. & Zimm. et al. \\ 
\hhline{-:=:=:=:=:=:=}
3D distance & 51.87 & 45.58 & \textbf{45.33} & 56.59 & 56.92 & 52.77\\
\hline 
\end{tabular}
}
\vspace{-5pt}
\caption{Average 3D joint distance (mm) to ground-truth for \textsc{EgoDexter}.}
\label{tab:ED}
\end{table} 
 
Figure \ref{fig:pck_DO} shows the performance of our method under occlusions and clutter with 3D PCK on the \textsc{Dexter+Object} dataset, and Table \ref{tab:D+O} shows 3D joint errors. Additionally, Figure \ref{fig:pck_ED} shows our results on a hand in ego-centric view and in interaction with various objects in terms of 3D PCK on the \textsc{EgoDexter} dataset, and Table \ref{tab:ED} shows 3D joint errors. Our method outperforms the competition in these settings as  illustrated in the Figures. We note that we show relative 3D pose estimates for all methods except \cite{iqbal2018hand} where the authors report absolute values.

\begin{figure}[h!]
\center
\includegraphics[width=.8\linewidth]{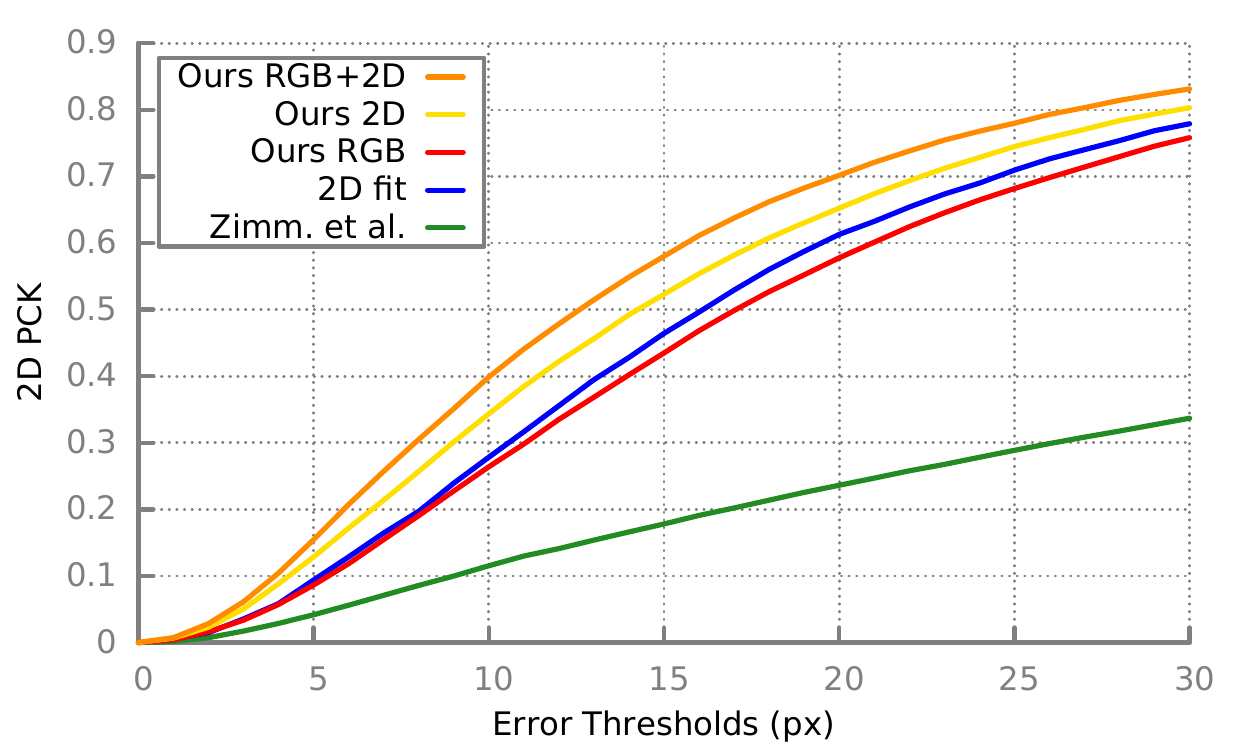}
\vspace{-8pt}
\caption{2D PCK for \textsc{Mpii+Nzsl}.}
\label{fig:pck_wild}
\end{figure}

\vspace{-5pt}
\begin{table}[h!]
\center
\resizebox{0.86\linewidth}{!}{%
\begin{tabular}{|c|*{5}{>{\centering\arraybackslash}m{1.0cm}|}}
\cline{2-6}
\multicolumn{1}{c|}{} & Ours {\footnotesize RGB} & Ours {\footnotesize RGB+2D} & Ours {\footnotesize 2D} & 2D fit & Zimm. et al. \\ 
\hhline{-:=:=:=:=:=}
2D distance & 23.04 & \textbf{18.95} & 20.65 & 22.36 & 59.40\\
\hline 
\end{tabular}
}
\vspace{-5pt}
\caption{Average re-projected 2D joint distance (px) to ground-truth for \textsc{Mpii+Nzsl}}
\label{tab:wild}
\end{table}

\begin{figure*}[t]
\center

\begin{subfigure}[c]{\sz\linewidth}
\includegraphics[width=\linewidth]{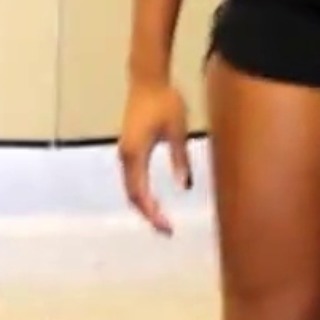}
\end{subfigure}
\begin{subfigure}[c]{\sz\linewidth}
\includegraphics[width=\linewidth]{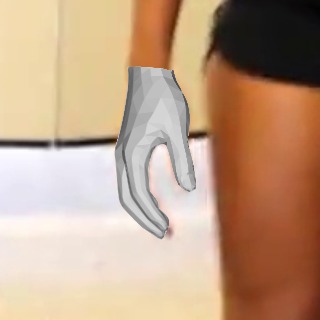}
\end{subfigure}
\begin{subfigure}[c]{\sz\linewidth}
\includegraphics[width=\linewidth]{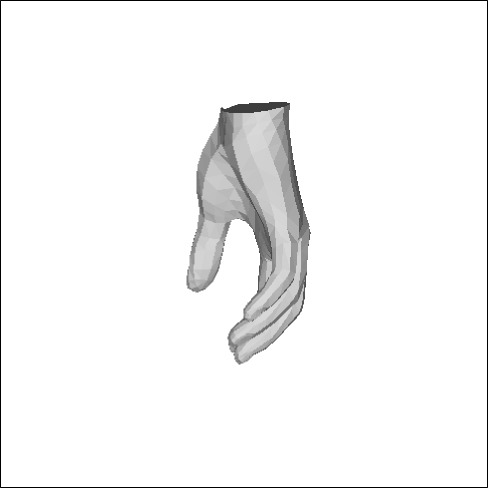}
\end{subfigure}
\begin{subfigure}[c]{\sz\linewidth}
\includegraphics[width=\linewidth]{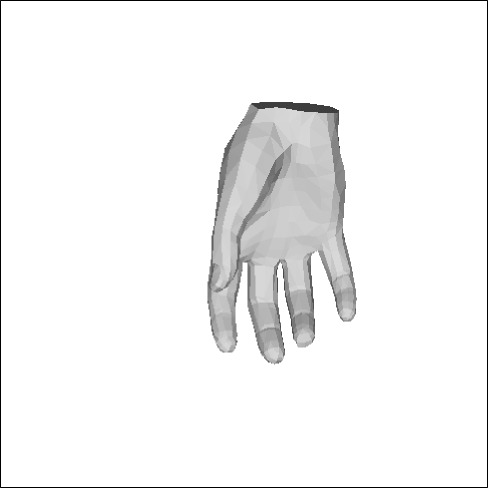}
\end{subfigure}
\begin{subfigure}[c]{\sz\linewidth}
\includegraphics[width=\linewidth]{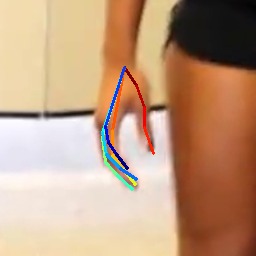}
\end{subfigure}
\hspace{3pt}
\begin{subfigure}[c]{\sz\linewidth}
\includegraphics[width=\linewidth]{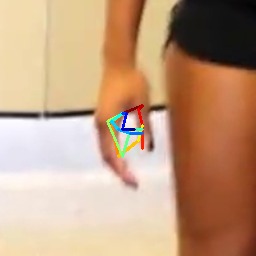}
\end{subfigure}
\begin{subfigure}[c]{\szl\linewidth}
\includegraphics[width=\linewidth]{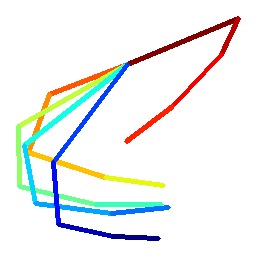}
\end{subfigure}
\hspace{3pt}
\begin{subfigure}[c]{\szl\linewidth}
\includegraphics[width=\linewidth]{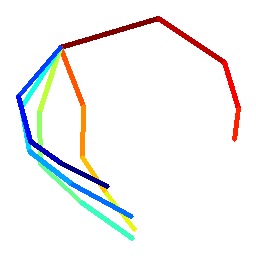}
\end{subfigure}

\begin{subfigure}[c]{\sz\linewidth}
\includegraphics[width=\linewidth]{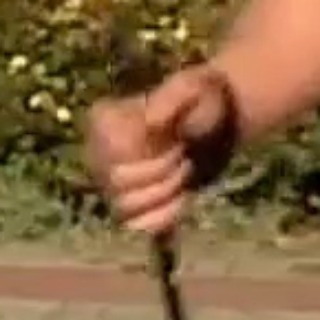}
\end{subfigure}
\begin{subfigure}[c]{\sz\linewidth}
\includegraphics[width=\linewidth]{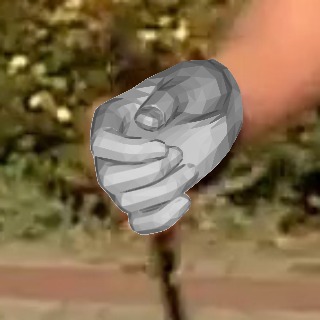}
\end{subfigure}
\begin{subfigure}[c]{\sz\linewidth}
\includegraphics[width=\linewidth]{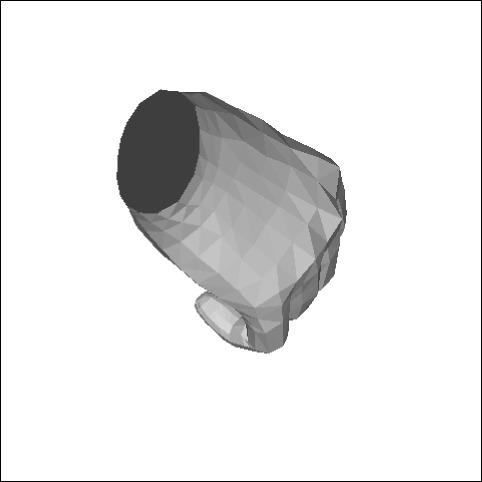}
\end{subfigure}
\begin{subfigure}[c]{\sz\linewidth}
\includegraphics[width=\linewidth]{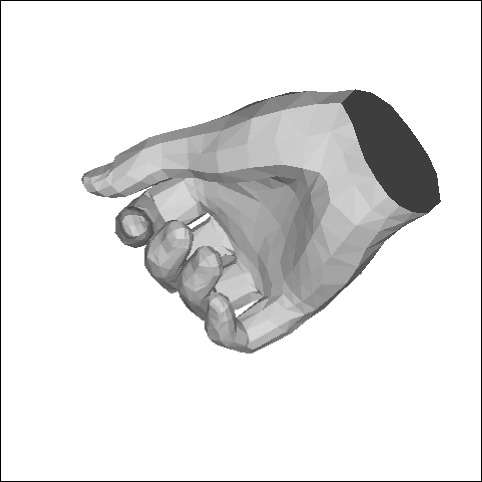}
\end{subfigure}
\begin{subfigure}[c]{\sz\linewidth}
\includegraphics[width=\linewidth]{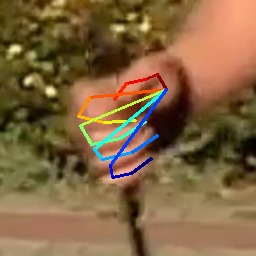}
\end{subfigure}
\hspace{3pt}
\begin{subfigure}[c]{\sz\linewidth}
\includegraphics[width=\linewidth]{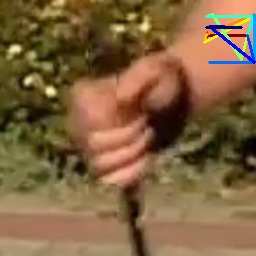}
\end{subfigure}
\begin{subfigure}[c]{\szl\linewidth}
\includegraphics[width=\linewidth]{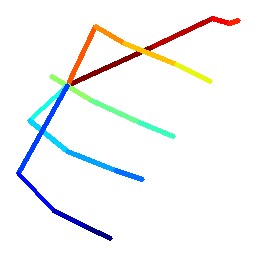}
\end{subfigure}
\hspace{3pt}
\begin{subfigure}[c]{\szl\linewidth}
\includegraphics[width=\linewidth]{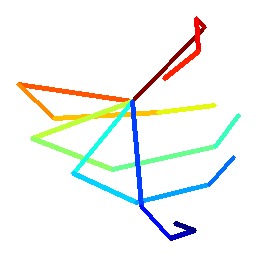}
\end{subfigure}

\begin{subfigure}[c]{\sz\linewidth}
\includegraphics[width=\linewidth]{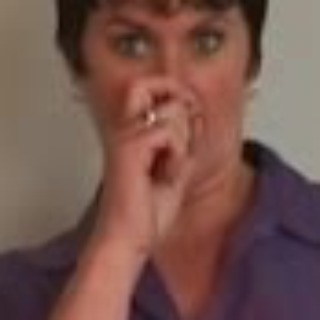}
\end{subfigure}
\begin{subfigure}[c]{\sz\linewidth}
\includegraphics[width=\linewidth]{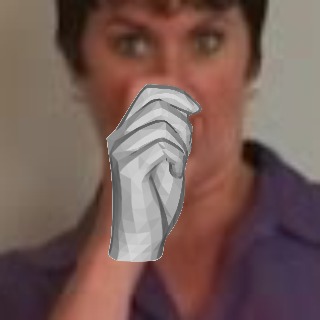}
\end{subfigure}
\begin{subfigure}[c]{\sz\linewidth}
\includegraphics[width=\linewidth]{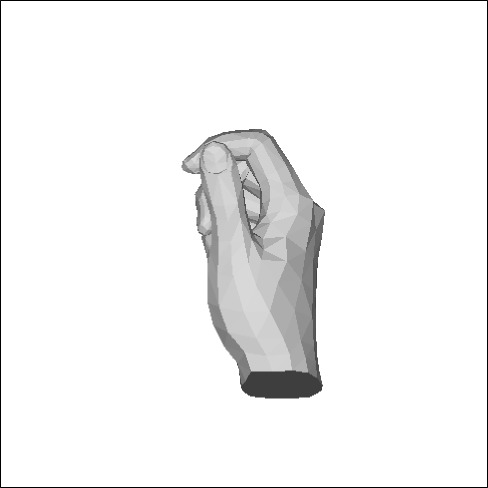}
\end{subfigure}
\begin{subfigure}[c]{\sz\linewidth}
\includegraphics[width=\linewidth]{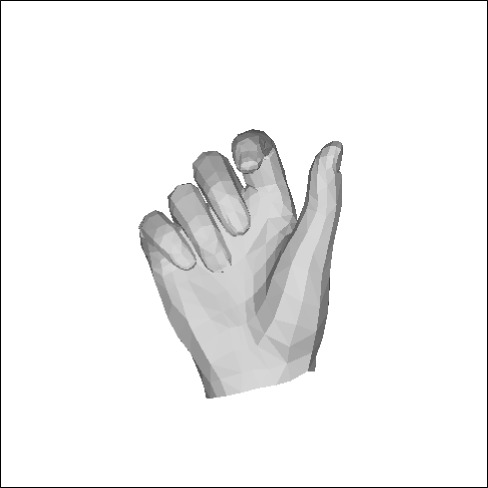}
\end{subfigure}
\begin{subfigure}[c]{\sz\linewidth}
\includegraphics[width=\linewidth]{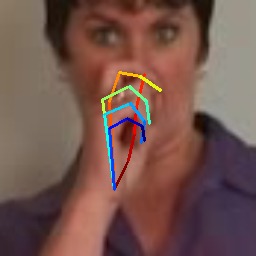}
\end{subfigure}
\hspace{3pt}
\begin{subfigure}[c]{\sz\linewidth}
\includegraphics[width=\linewidth]{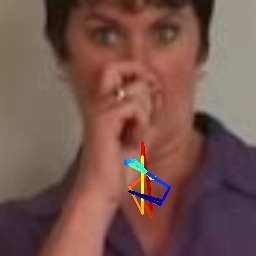}
\end{subfigure}
\begin{subfigure}[c]{\szl\linewidth}
\includegraphics[width=\linewidth]{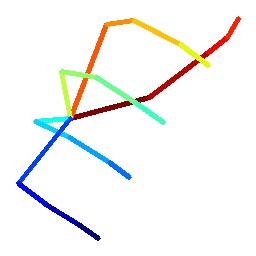}
\end{subfigure}
\hspace{3pt}
\begin{subfigure}[c]{\szl\linewidth}
\includegraphics[width=\linewidth]{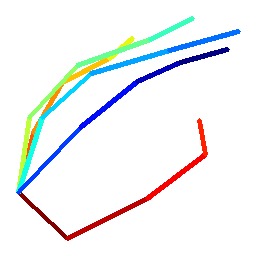}
\end{subfigure}

\begin{subfigure}[c]{\sz\linewidth}
\includegraphics[width=\linewidth]{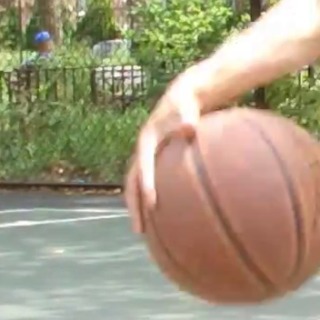}
\end{subfigure}
\begin{subfigure}[c]{\sz\linewidth}
\includegraphics[width=\linewidth]{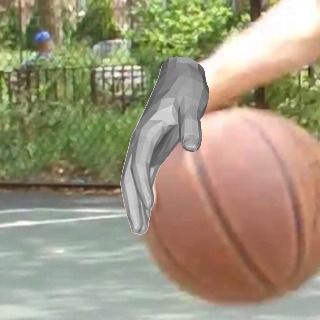}
\end{subfigure}
\begin{subfigure}[c]{\sz\linewidth}
\includegraphics[width=\linewidth]{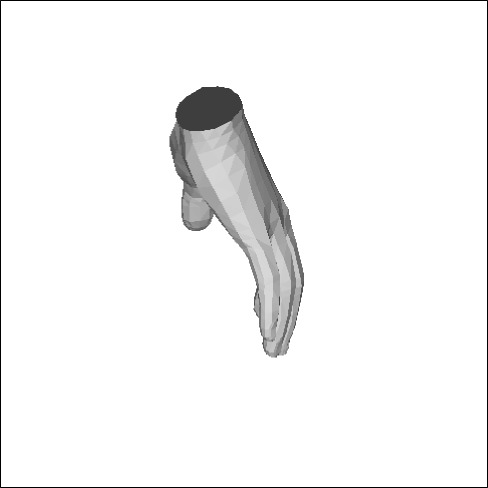}
\end{subfigure}
\begin{subfigure}[c]{\sz\linewidth}
\includegraphics[width=\linewidth]{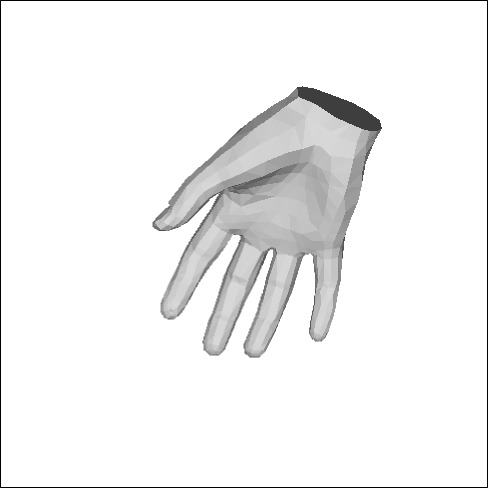}
\end{subfigure}
\begin{subfigure}[c]{\sz\linewidth}
\includegraphics[width=\linewidth]{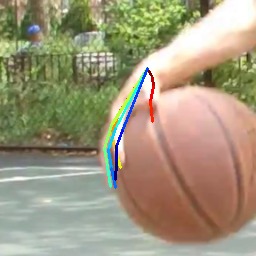}
\end{subfigure}
\hspace{3pt}
\begin{subfigure}[c]{\sz\linewidth}
\includegraphics[width=\linewidth]{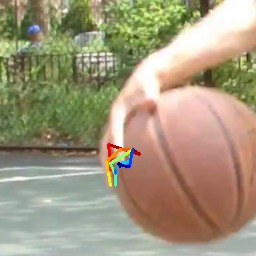}
\end{subfigure}
\begin{subfigure}[c]{\szl\linewidth}
\includegraphics[width=\linewidth]{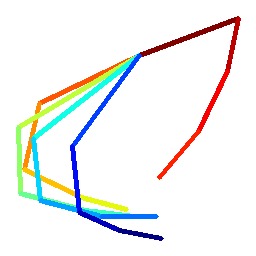}
\end{subfigure}
\hspace{3pt}
\begin{subfigure}[c]{\szl\linewidth}
\includegraphics[width=\linewidth]{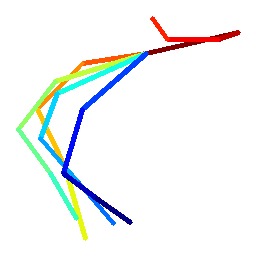}
\end{subfigure}

\begin{subfigure}[c]{\sz\linewidth}
\includegraphics[width=\linewidth]{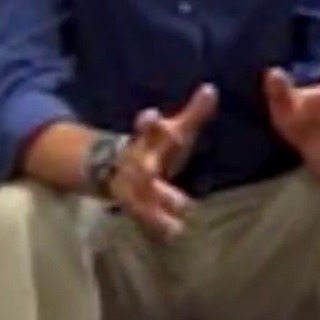}
\end{subfigure}
\begin{subfigure}[c]{\sz\linewidth}
\includegraphics[width=\linewidth]{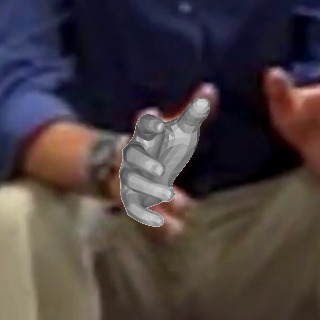}
\end{subfigure}
\begin{subfigure}[c]{\sz\linewidth}
\includegraphics[width=\linewidth]{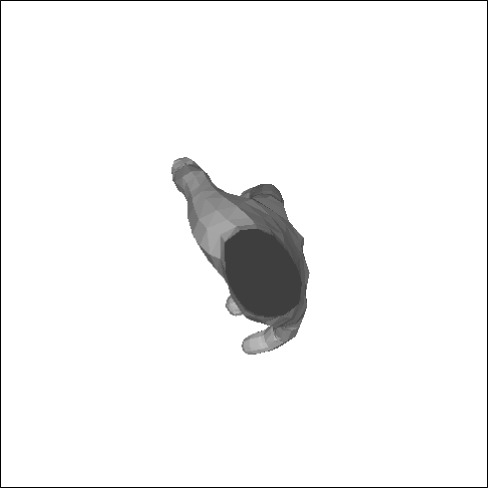}
\end{subfigure}
\begin{subfigure}[c]{\sz\linewidth}
\includegraphics[width=\linewidth]{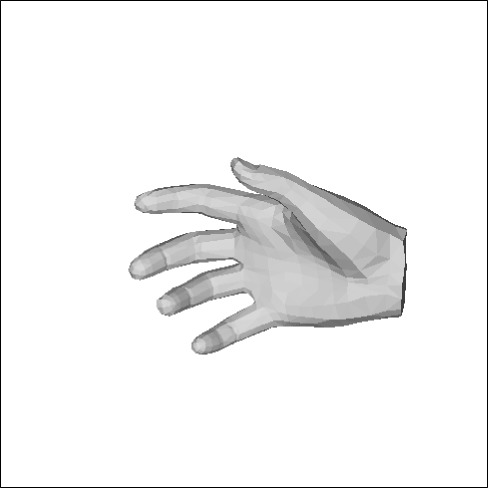}
\end{subfigure}
\begin{subfigure}[c]{\sz\linewidth}
\includegraphics[width=\linewidth]{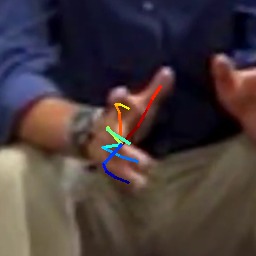}
\end{subfigure}
\hspace{3pt}
\begin{subfigure}[c]{\sz\linewidth}
\includegraphics[width=\linewidth]{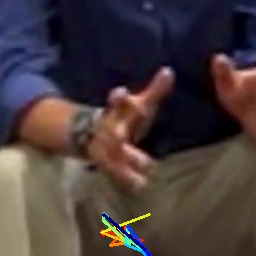}
\end{subfigure}
\begin{subfigure}[c]{\szl\linewidth}
\includegraphics[width=\linewidth]{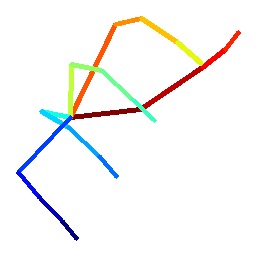}
\end{subfigure}
\hspace{3pt}
\begin{subfigure}[c]{\szl\linewidth}
\includegraphics[width=\linewidth]{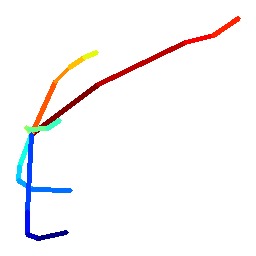}
\end{subfigure}

\begin{subfigure}[c]{\sz\linewidth}
\includegraphics[width=\linewidth]{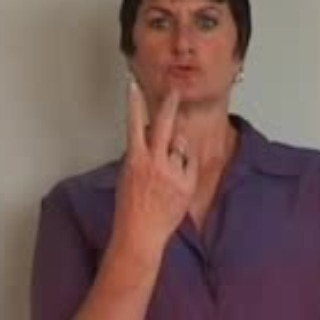}
\end{subfigure}
\begin{subfigure}[c]{\sz\linewidth}
\includegraphics[width=\linewidth]{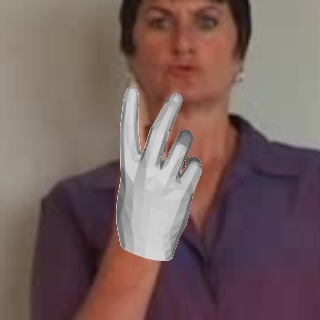}
\end{subfigure}
\begin{subfigure}[c]{\sz\linewidth}
\includegraphics[width=\linewidth]{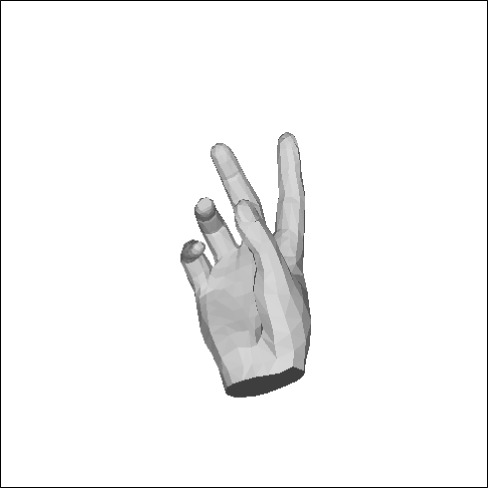}
\end{subfigure}
\begin{subfigure}[c]{\sz\linewidth}
\includegraphics[width=\linewidth]{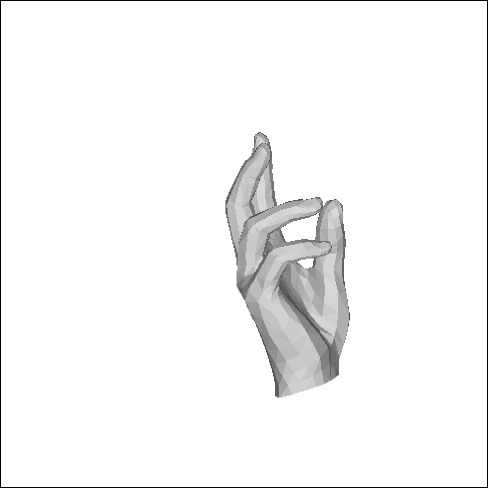}
\end{subfigure}
\begin{subfigure}[c]{\sz\linewidth}
\includegraphics[width=\linewidth]{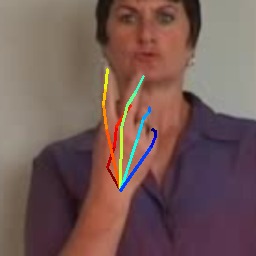}
\end{subfigure}
\hspace{3pt}
\begin{subfigure}[c]{\sz\linewidth}
\includegraphics[width=\linewidth]{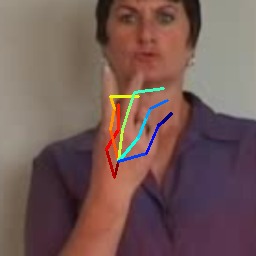}
\end{subfigure}
\begin{subfigure}[c]{\szl\linewidth}
\includegraphics[width=\linewidth]{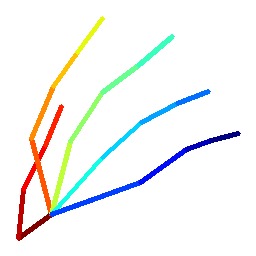}
\end{subfigure}
\hspace{3pt}
\begin{subfigure}[c]{\szl\linewidth}
\includegraphics[width=\linewidth]{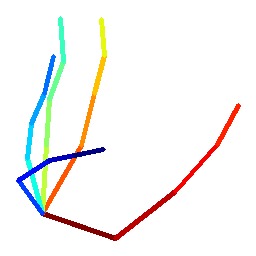}
\end{subfigure}

\begin{subfigure}[c]{\sz\linewidth}
\includegraphics[width=\linewidth]{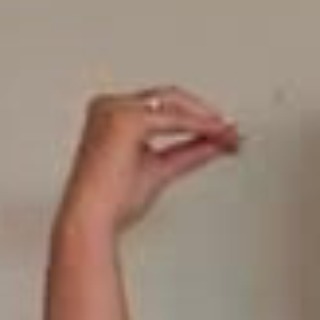}
\end{subfigure}
\begin{subfigure}[c]{\sz\linewidth}
\includegraphics[width=\linewidth]{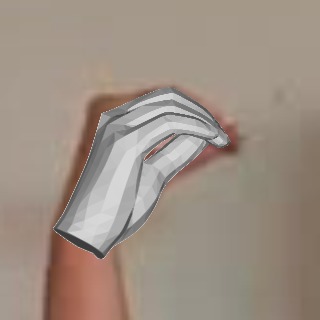}
\end{subfigure}
\begin{subfigure}[c]{\sz\linewidth}
\includegraphics[width=\linewidth]{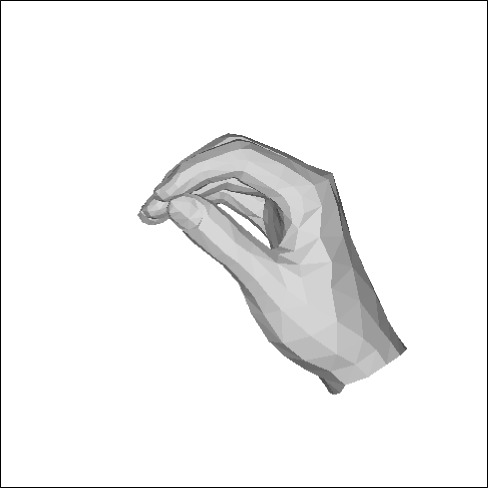}
\end{subfigure}
\begin{subfigure}[c]{\sz\linewidth}
\includegraphics[width=\linewidth]{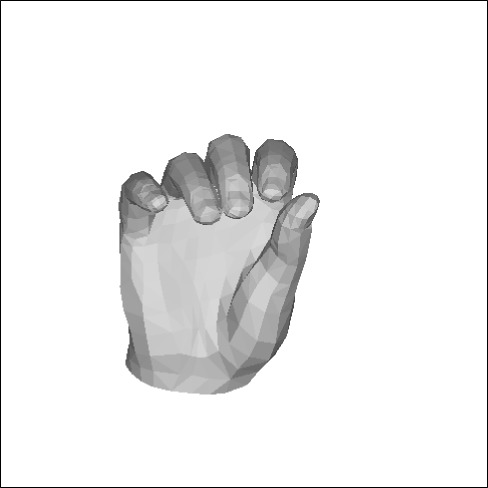}
\end{subfigure}
\begin{subfigure}[c]{\sz\linewidth}
\includegraphics[width=\linewidth]{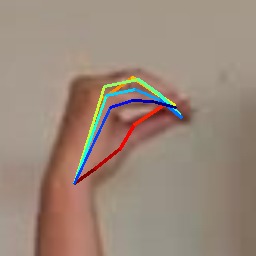}
\end{subfigure}
\hspace{3pt}
\begin{subfigure}[c]{\sz\linewidth}
\includegraphics[width=\linewidth]{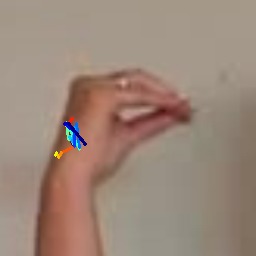}
\end{subfigure}
\begin{subfigure}[c]{\szl\linewidth}
\includegraphics[width=\linewidth]{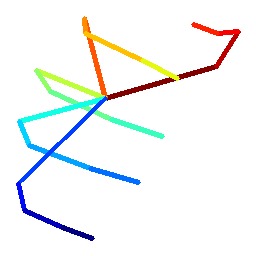}
\end{subfigure}
\hspace{3pt}
\begin{subfigure}[c]{\szl\linewidth}
\includegraphics[width=\linewidth]{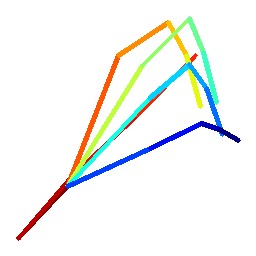}
\end{subfigure}

\begin{subfigure}[c]{\sz\linewidth}
\includegraphics[width=\linewidth]{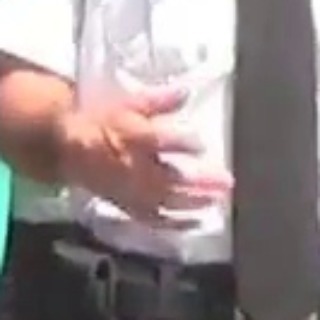}
\end{subfigure}
\begin{subfigure}[c]{\sz\linewidth}
\includegraphics[width=\linewidth]{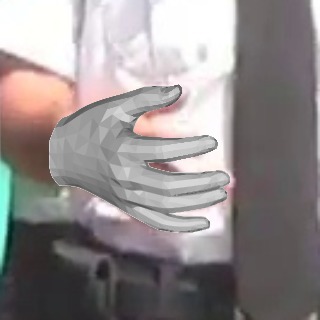}
\end{subfigure}
\begin{subfigure}[c]{\sz\linewidth}
\includegraphics[width=\linewidth]{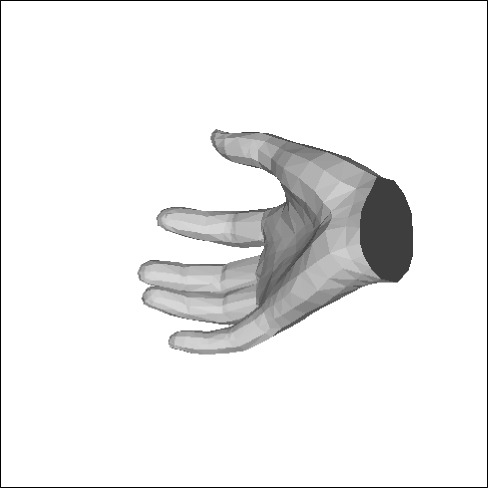}
\end{subfigure}
\begin{subfigure}[c]{\sz\linewidth}
\includegraphics[width=\linewidth]{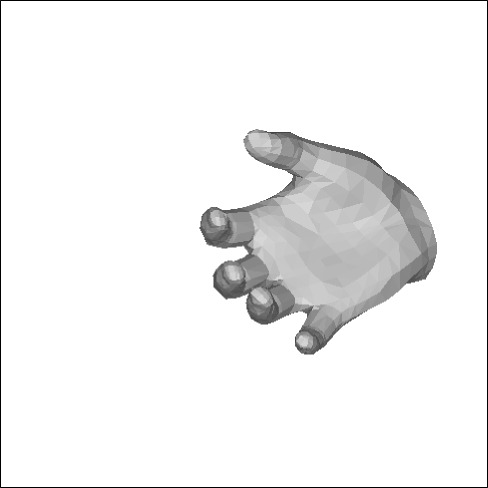}
\end{subfigure}
\begin{subfigure}[c]{\sz\linewidth}
\includegraphics[width=\linewidth]{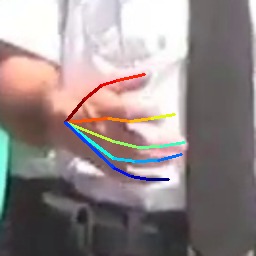}
\end{subfigure}
\hspace{3pt}
\begin{subfigure}[c]{\sz\linewidth}
\includegraphics[width=\linewidth]{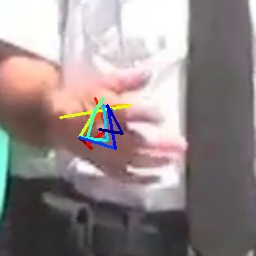}
\end{subfigure}
\begin{subfigure}[c]{\szl\linewidth}
\includegraphics[width=\linewidth]{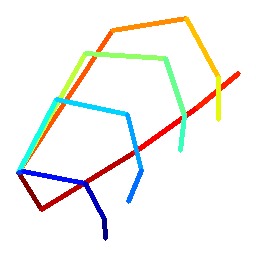}
\end{subfigure}
\hspace{3pt}
\begin{subfigure}[c]{\szl\linewidth}
\includegraphics[width=\linewidth]{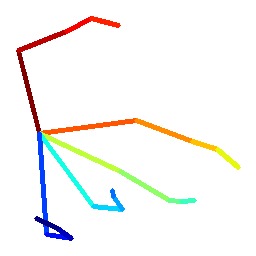}
\end{subfigure}

\begin{subfigure}[c]{\sz\linewidth}
\includegraphics[width=\linewidth]{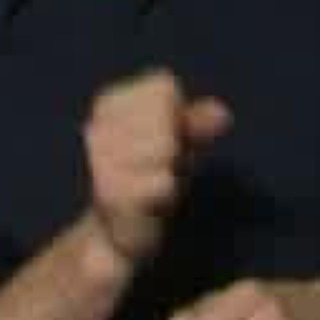}
\end{subfigure}
\begin{subfigure}[c]{\sz\linewidth}
\includegraphics[width=\linewidth]{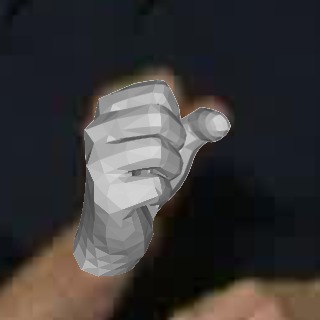}
\end{subfigure}
\begin{subfigure}[c]{\sz\linewidth}
\includegraphics[width=\linewidth]{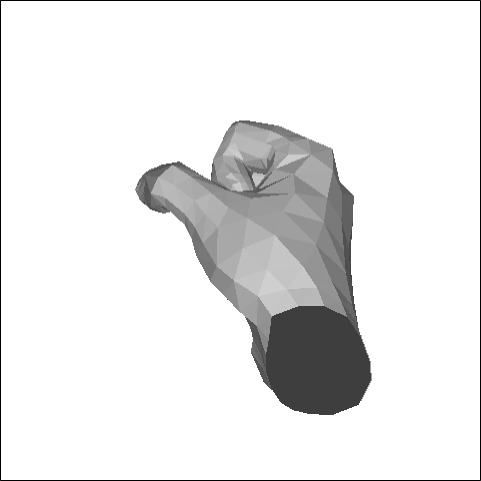}
\end{subfigure}
\begin{subfigure}[c]{\sz\linewidth}
\includegraphics[width=\linewidth]{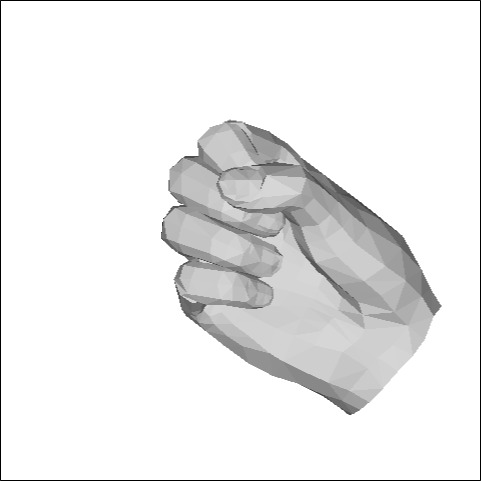}
\end{subfigure}
\begin{subfigure}[c]{\sz\linewidth}
\includegraphics[width=\linewidth]{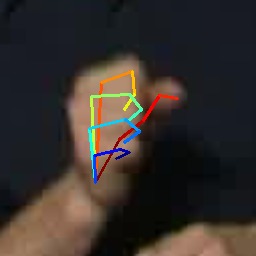}
\end{subfigure}
\hspace{3pt}
\begin{subfigure}[c]{\sz\linewidth}
\includegraphics[width=\linewidth]{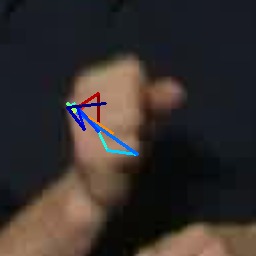}
\end{subfigure}
\begin{subfigure}[c]{\szl\linewidth}
\includegraphics[width=\linewidth]{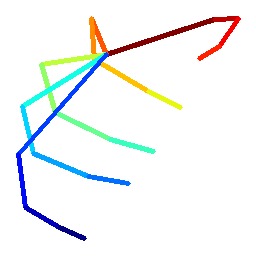}
\end{subfigure}
\hspace{3pt}
\begin{subfigure}[c]{\szl\linewidth}
\includegraphics[width=\linewidth]{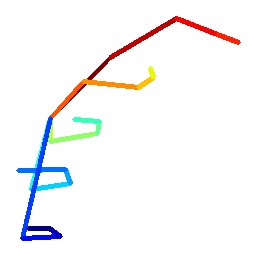}
\end{subfigure}

\begin{subfigure}[c]{\sz\linewidth}
\includegraphics[width=\linewidth]{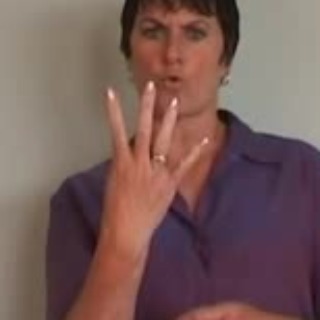}
\end{subfigure}
\begin{subfigure}[c]{\sz\linewidth}
\includegraphics[width=\linewidth]{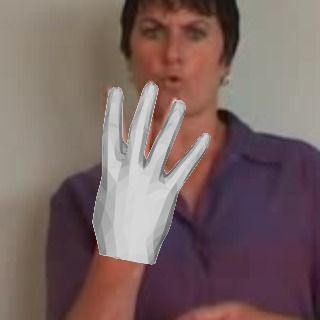}
\end{subfigure}
\begin{subfigure}[c]{\sz\linewidth}
\includegraphics[width=\linewidth]{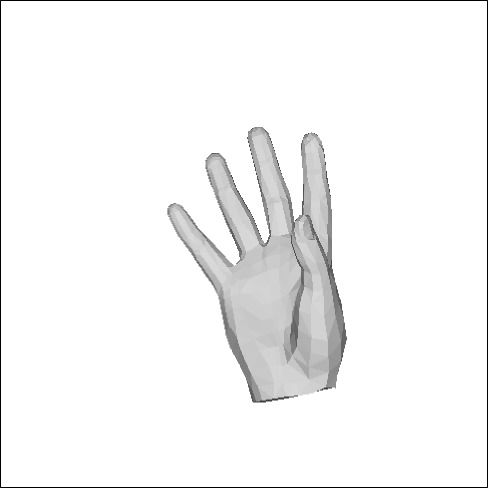}
\end{subfigure}
\begin{subfigure}[c]{\sz\linewidth}
\includegraphics[width=\linewidth]{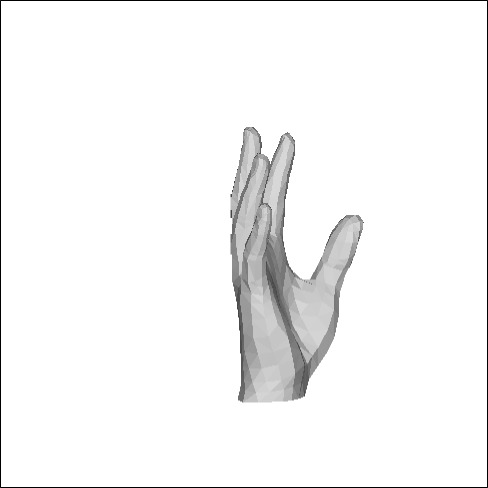}
\end{subfigure}
\begin{subfigure}[c]{\sz\linewidth}
\includegraphics[width=\linewidth]{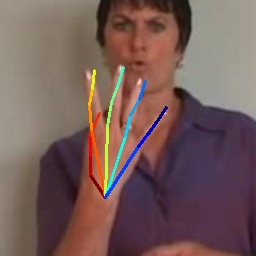}
\end{subfigure}
\hspace{3pt}
\begin{subfigure}[c]{\sz\linewidth}
\includegraphics[width=\linewidth]{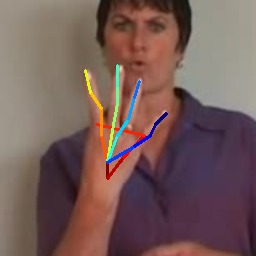}
\end{subfigure}
\begin{subfigure}[c]{\szl\linewidth}
\includegraphics[width=\linewidth]{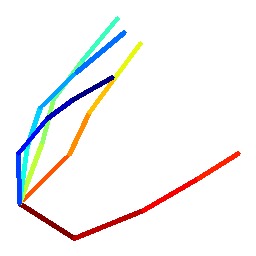}
\end{subfigure}
\hspace{3pt}
\begin{subfigure}[c]{\szl\linewidth}
\includegraphics[width=\linewidth]{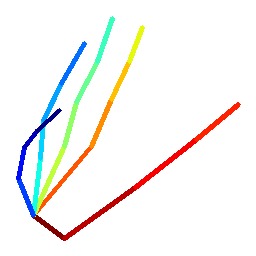}
\end{subfigure}

\begin{subfigure}[c]{\sz\linewidth}
\includegraphics[width=\linewidth]{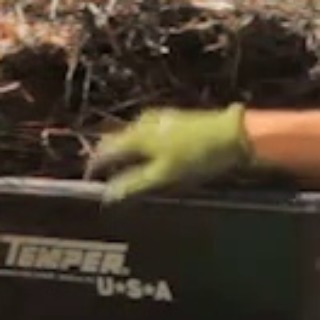}
\end{subfigure}
\begin{subfigure}[c]{\sz\linewidth}
\includegraphics[width=\linewidth]{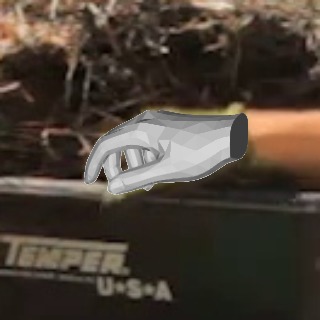}
\end{subfigure}
\begin{subfigure}[c]{\sz\linewidth}
\includegraphics[width=\linewidth]{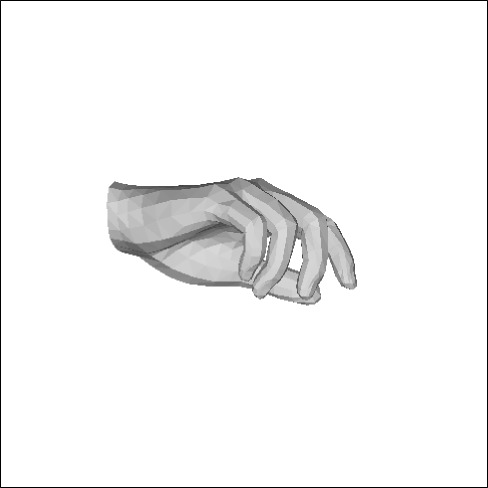}
\end{subfigure}
\begin{subfigure}[c]{\sz\linewidth}
\includegraphics[width=\linewidth]{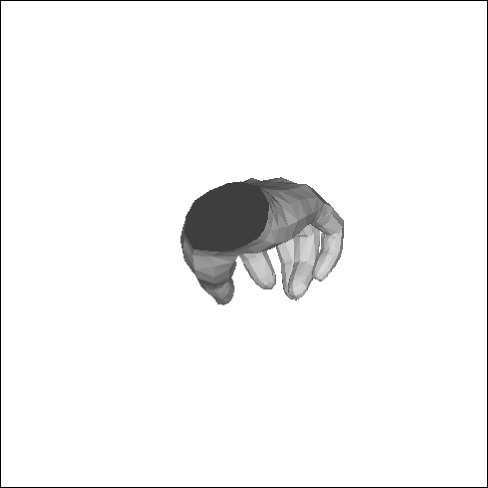}
\end{subfigure}
\begin{subfigure}[c]{\sz\linewidth}
\includegraphics[width=\linewidth]{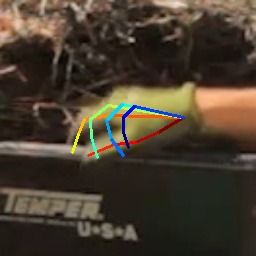}
\end{subfigure}
\hspace{3pt}
\begin{subfigure}[c]{\sz\linewidth}
\includegraphics[width=\linewidth]{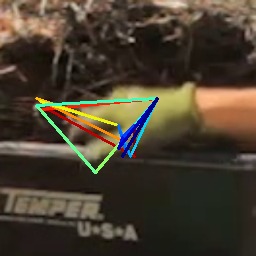}
\end{subfigure}
\begin{subfigure}[c]{\szl\linewidth}
\includegraphics[width=\linewidth]{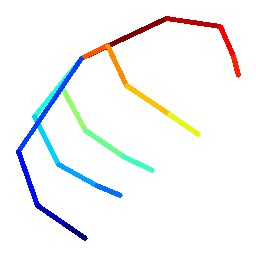}
\end{subfigure}
\hspace{3pt}
\begin{subfigure}[c]{\szl\linewidth}
\includegraphics[width=\linewidth]{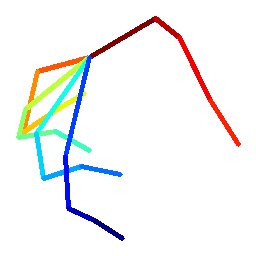}
\end{subfigure}

\begin{subfigure}[c]{\sz\linewidth}
\includegraphics[width=\linewidth]{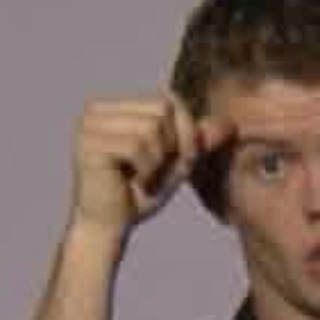}
\caption{\scriptsize Input}
\label{fig:qual_input}
\end{subfigure}
\begin{subfigure}[c]{\sz\linewidth}
\includegraphics[width=\linewidth]{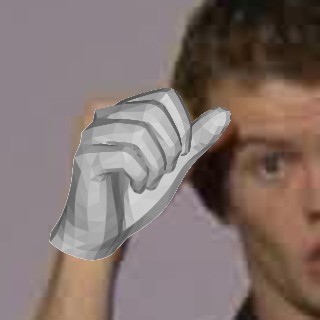}
\caption{\scriptsize Our mesh}
\label{fig:qual_o1}
\end{subfigure}
\begin{subfigure}[c]{\sz\linewidth}
\includegraphics[width=\linewidth]{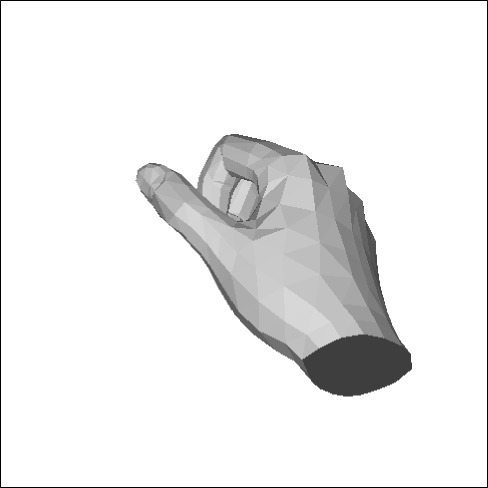}
\caption{\scriptsize Back view}
\label{fig:qual_o2}
\end{subfigure}
\begin{subfigure}[c]{\sz\linewidth}
\includegraphics[width=\linewidth]{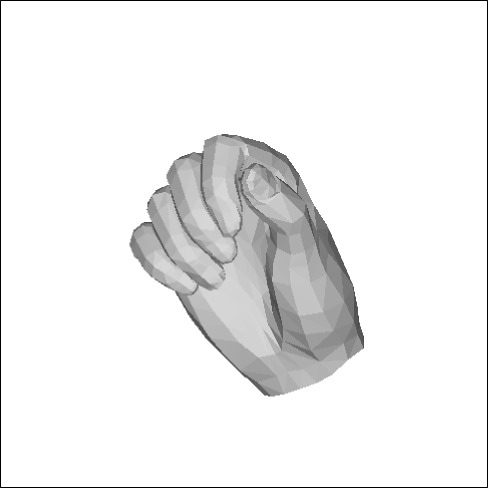}
\caption{\scriptsize Side view}
\label{fig:qual_o3}
\end{subfigure}
\begin{subfigure}[c]{\sz\linewidth}
\includegraphics[width=\linewidth]{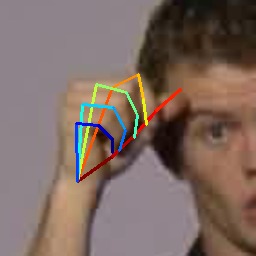}
\caption{\scriptsize Our skeleton}
\label{fig:qual_o4}
\end{subfigure}
\hspace{3pt}
\begin{subfigure}[c]{\sz\linewidth}
\includegraphics[width=\linewidth]{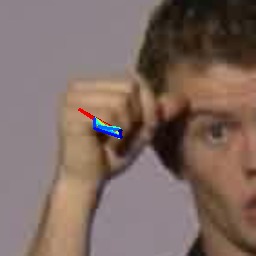}
\caption{\scriptsize\scriptsize\cite{zimmermann2017learning}2D}
\label{fig:qual_z2}
\end{subfigure}
\begin{subfigure}[c]{\szl\linewidth}
\includegraphics[width=\linewidth]{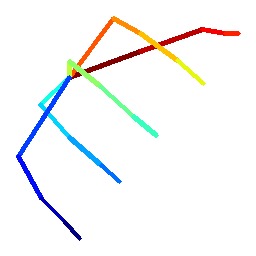}
\caption{\scriptsize\cite{zimmermann2017learning}3D}
\label{fig:qual_z3}
\end{subfigure}
\hspace{3pt}
\begin{subfigure}[c]{\szl\linewidth}
\includegraphics[width=\linewidth]{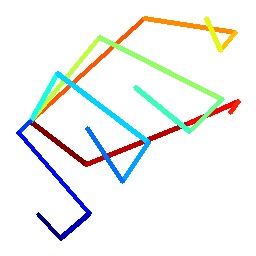}
\caption{\scriptsize\cite{spurr2018cross}}
\label{fig:qual_s}
\end{subfigure}

\vspace{-7pt}

\caption{Our 3D hand reconstruction on examples from the challenging testing set of \textsc{Mpii+Nzsl} compared to the 3D hand pose predictions of \cite{zimmermann2017learning} and \cite{spurr2018cross}.}
\label{fig:qual}

\vspace{\szm}

\end{figure*}

We expect our method to perform particularly well on datasets of images in the wild, as our training set contains this type of data and accounts for hands in low resolution, blurry, occluded and in challenging views and pose configurations. In fact, we compare our results to \cite{zimmermann2017learning} on the testing set of \textsc{Mpii+Nzsl} dataset in Figure \ref{fig:pck_wild} and Table \ref{tab:wild} through 2D PCK and 2D joint error respectively. We outperform \cite{zimmermann2017learning} with a substantial margin as the Figure shows. The superiority of our method on this dataset is visually confirmed in Figure \ref{fig:qual}.

\paragraph{Comparison to 2D fitting}
\vspace{-10pt}
In the case where 2D joint detections are used as input, an alternative way of solving 3D hand pose estimation is to perform a 2D fitting between the re-projected hand model joints and the key-points detected on the image, in a similar fashion to the work proposed by \cite{panteleris2018using}. Our implementation of this strategy consists in minimizing the following objective function with respect to the weak perspective camera parameters $\{R,t,s\}$ and the hand shape and pose parameters $\{\boldsymbol{\beta},\boldsymbol{\theta}\}$:
\begin{equation}
\begin{aligned}
E(R,t,s,\boldsymbol{\beta},\boldsymbol{\theta}) = &\sum_i p_i \left(s\Pi(RJ_i(\boldsymbol{\beta},\boldsymbol{\theta}))+t - \mathrm{x}_i \right)^2\\ 
& + \alpha_{\boldsymbol{\beta}} \|\boldsymbol{\beta}\|_2^2 + \|\boldsymbol{\theta}\|_2^2,
\end{aligned}
\label{eq:opt}
\end{equation}
where $p_i$ is the $i^{th}$ hand joint estimate confidence provided by the detector CNN \cite{simon2017hand}. Similarly to the loss in Equation \ref{eq:reg}, regularization in the second line of Equation \ref{eq:opt} is important to ensure plausible 3D hand reconstructions. We perform this optimization using Powell's Dogleg method \cite{nocedal2006nonlinear} within the Chumpy \cite{chumpy} framework. 

We compare this method (2D fit) to our proposed approach on datasets \textsc{Stereo}, \textsc{Dexter+Object} and \textsc{EgoDexter} with 3D PCK in Figures \ref{fig:pck_s2}, \ref{fig:pck_DO} and \ref{fig:pck_ED} and 3D joint error in Tables \ref{tab:Stereo}, \ref{tab:D+O} and \ref{tab:ED} respectively, and also on dataset \textsc{Mpii+Nzsl} with 2D PCK in Figure \ref{fig:pck_wild} and 2D joint error in Table \ref{tab:wild}. Results show that our approach outperforms the 2D fitting based strategy for all datasets. We observe that while the optimization catches up slightly with our method in 2D (\textsc{Mpii+Nzsl}), its performance drops considerably in 3D. Our method benefits clearly from solving the fitting problem in a learning framework and leverages visual cues in predicting the 3D hand position and configuration, while the 2D fitting relies merely on the 2D joint detection information. We also outperform the 2D fitting based method in \cite{panteleris2018using} that uses a similar hand model to \cite{oikonomidis2011efficient} and a perspective projection model on dataset \textsc{Stereo} in Figure \ref{fig:pck_s2}. 
 
\paragraph{Ablation study}
\vspace{-10pt}
We evaluate the difference between using images only (Ours RGB), using 2D joint heat-maps obtained from a state-of-the-art hand detector \cite{simon2017hand} only (Ours 2D), and finally using both together as input (Ours RGB+2D). We carry comparisons on datasets \textsc{Stereo}, \textsc{Dexter+Object} and \textsc{EgoDexter} with 3D PCK in Figures \ref{fig:pck_s2}, \ref{fig:pck_DO} and \ref{fig:pck_ED} and 3D joint error in Tables \ref{tab:Stereo}, \ref{tab:D+O} and \ref{tab:ED} respectively, and also on dataset \textsc{Mpii+Nzsl} with 2D PCK in Figure \ref{fig:pck_wild} and 2D joint error in Table \ref{tab:wild}. On dataset \textsc{Stereo}, training on images alone yields the best performance, while training with a combination of images and 2D joint heat-maps is generally the most suitable approach for the other datasets that we tested on. 

\paragraph{Qualitative}
\vspace{-10pt}
Figure \ref{fig:qual} shows our 3D hand reconstructions on the challenging testing set of \textsc{Mpii+Nzsl}. As shown in this Figure, the input data (\ref{fig:qual_input}) displays images of hands that are sometimes blurry, low resolved, occluded, viewed from varying viewpoints and in varying pose configurations. We show our 3D mesh overlaid on the input image (\ref{fig:qual_o1}) and in alternative views (\ref{fig:qual_o2}, \ref{fig:qual_o3}). We also compare our hand skeleton (\ref{fig:qual_o4}) to the 2D and 3D pose predictions of \cite{zimmermann2017learning} (\ref{fig:qual_z2}, \ref{fig:qual_z3}) and the 3D predictions of \cite{spurr2018cross} (\ref{fig:qual_s}). Our method obtains visually plausible results while the methods in \cite{zimmermann2017learning} and \cite{spurr2018cross} fail to predict good 3D pose estimates for many cases in the \textsc{Mpii+Nzsl} dataset. We show more examples in the supplementary material. 

\section{Conclusion}

We presented a method to predict 3D hand pose and shape from a single RGB image. We combine a deep convolutional encoder with a generative hand model as decoder and train the resulting network end-to-end with 2D and 3D hand joint annotated images. The encoder predicts hand parameters that are inputted to the hand model, and view parameters that are used to re-project the generated 3D hand into the image domain. We generate state-of-the-art results on 3D pose benchmarks and show compelling 3D reconstruction on a challenging set of images in the wild. This method could benefit greatly from a hand appearance model by leveraging a photometric loss in training as proposed in \cite{tewari2017mofa,tewari2017self} for faces. 
One possible extension to this work could be to allow some components of the MANO \cite{romero2017embodied} model such as the corrective blend shapes $\mathbf{S}$ and $\mathbf{P}$ (Equation \ref{eq:def}) to be fine-tuned in training for improved performance.   

{\small
\bibliographystyle{ieee}
\bibliography{egbib}
}

%
%
%


\begin{figure*}[h]
\center

\begin{subfigure}[c]{\sze\linewidth}
\includegraphics[width=\linewidth]{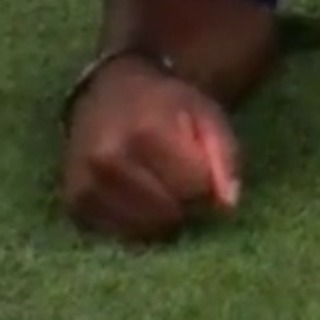}
\end{subfigure}
\begin{subfigure}[c]{\sze\linewidth}
\includegraphics[width=\linewidth]{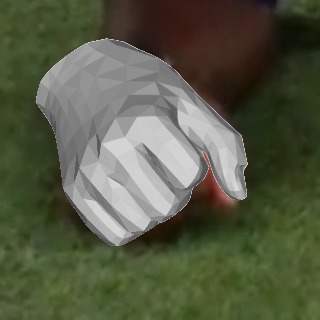}
\end{subfigure}
\begin{subfigure}[c]{\sze\linewidth}
\includegraphics[width=\linewidth]{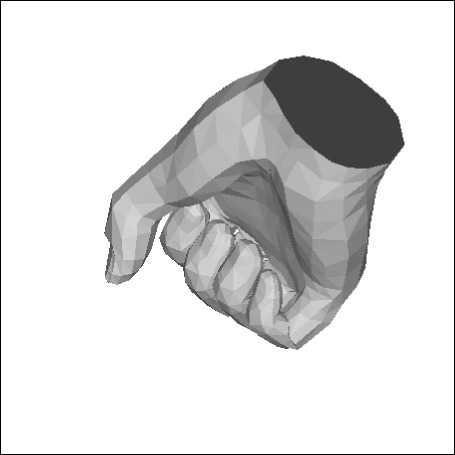}
\end{subfigure}
\begin{subfigure}[c]{\sze\linewidth}
\includegraphics[width=\linewidth]{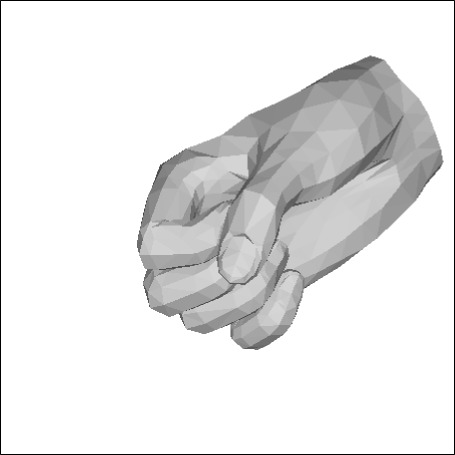}
\end{subfigure}
\begin{subfigure}[c]{\sze\linewidth}
\includegraphics[width=\linewidth]{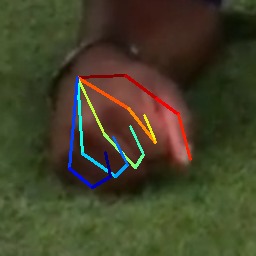}
\end{subfigure}
\hspace{3pt}
\begin{subfigure}[c]{\sze\linewidth}
\includegraphics[width=\linewidth]{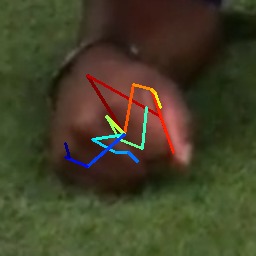}
\end{subfigure}
\begin{subfigure}[c]{\szle\linewidth}
\includegraphics[width=\linewidth]{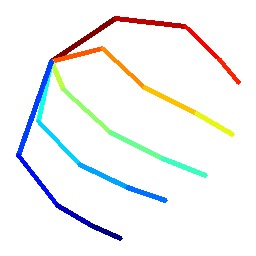}
\end{subfigure}
\hspace{3pt}
\begin{subfigure}[c]{\szle\linewidth}
\includegraphics[width=\linewidth]{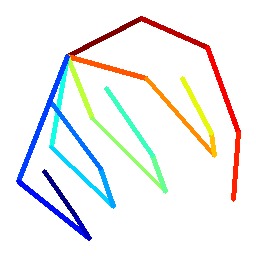}
\end{subfigure}

\begin{subfigure}[c]{\sze\linewidth}
\includegraphics[width=\linewidth]{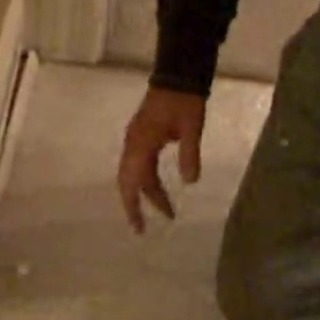}
\end{subfigure}
\begin{subfigure}[c]{\sze\linewidth}
\includegraphics[width=\linewidth]{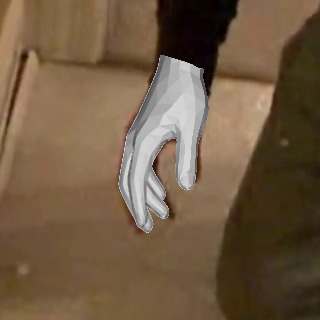}
\end{subfigure}
\begin{subfigure}[c]{\sze\linewidth}
\includegraphics[width=\linewidth]{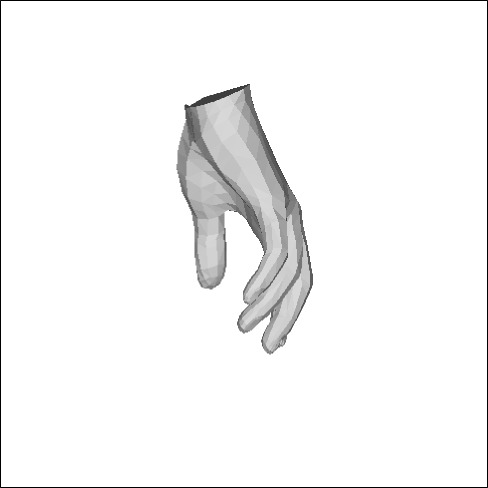}
\end{subfigure}
\begin{subfigure}[c]{\sze\linewidth}
\includegraphics[width=\linewidth]{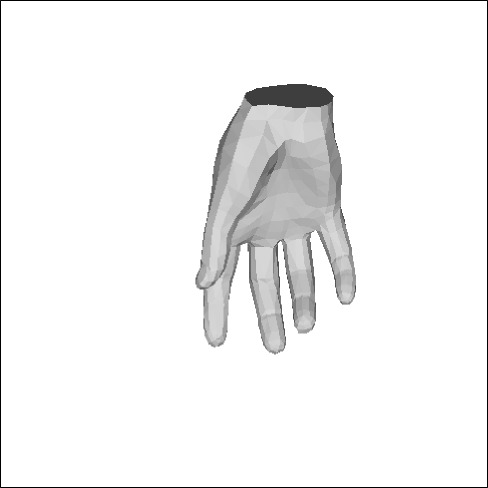}
\end{subfigure}
\begin{subfigure}[c]{\sze\linewidth}
\includegraphics[width=\linewidth]{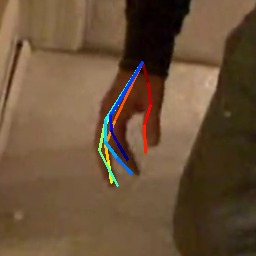}
\end{subfigure}
\hspace{3pt}
\begin{subfigure}[c]{\sze\linewidth}
\includegraphics[width=\linewidth]{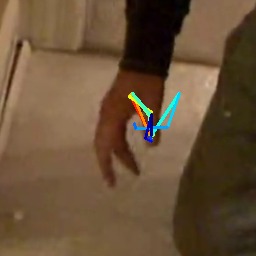}
\end{subfigure}
\begin{subfigure}[c]{\szle\linewidth}
\includegraphics[width=\linewidth]{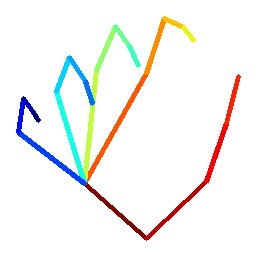}
\end{subfigure}
\hspace{3pt}
\begin{subfigure}[c]{\szle\linewidth}
\includegraphics[width=\linewidth]{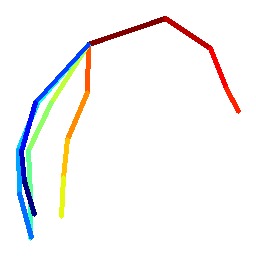}
\end{subfigure}

\begin{subfigure}[c]{\sze\linewidth}
\includegraphics[width=\linewidth]{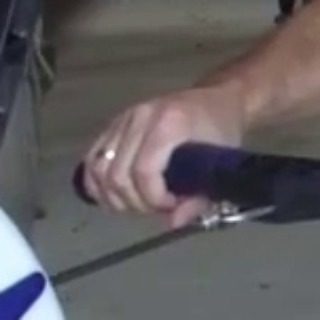}
\end{subfigure}
\begin{subfigure}[c]{\sze\linewidth}
\includegraphics[width=\linewidth]{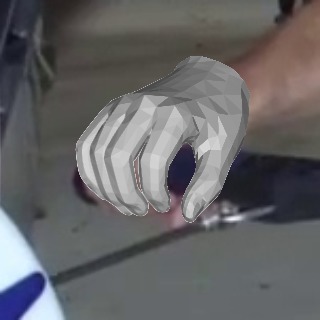}
\end{subfigure}
\begin{subfigure}[c]{\sze\linewidth}
\includegraphics[width=\linewidth]{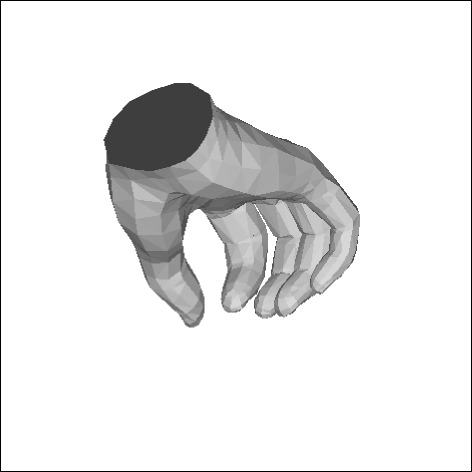}
\end{subfigure}
\begin{subfigure}[c]{\sze\linewidth}
\includegraphics[width=\linewidth]{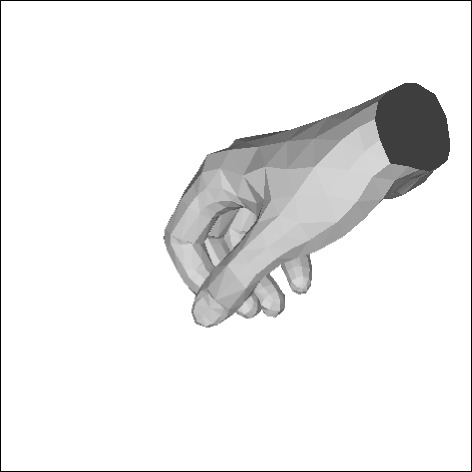}
\end{subfigure}
\begin{subfigure}[c]{\sze\linewidth}
\includegraphics[width=\linewidth]{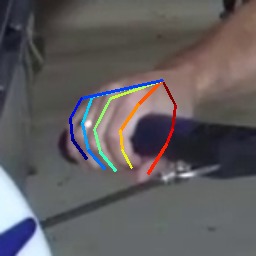}
\end{subfigure}
\hspace{3pt}
\begin{subfigure}[c]{\sze\linewidth}
\includegraphics[width=\linewidth]{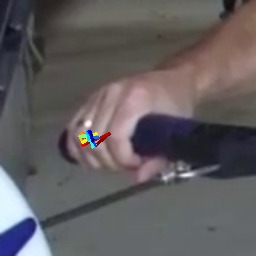}
\end{subfigure}
\begin{subfigure}[c]{\szle\linewidth}
\includegraphics[width=\linewidth]{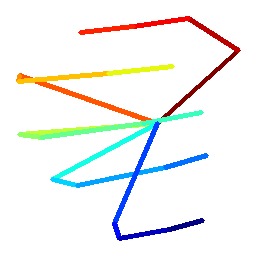}
\end{subfigure}
\hspace{3pt}
\begin{subfigure}[c]{\szle\linewidth}
\includegraphics[width=\linewidth]{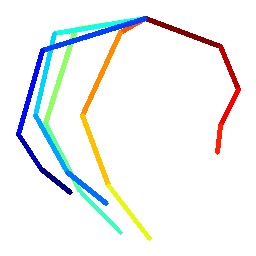}
\end{subfigure}

\begin{subfigure}[c]{\sze\linewidth}
\includegraphics[width=\linewidth]{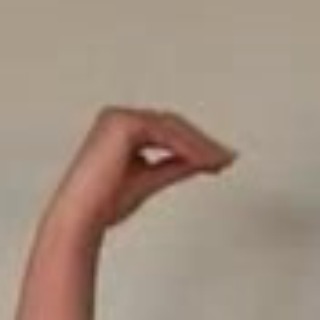}
\end{subfigure}
\begin{subfigure}[c]{\sze\linewidth}
\includegraphics[width=\linewidth]{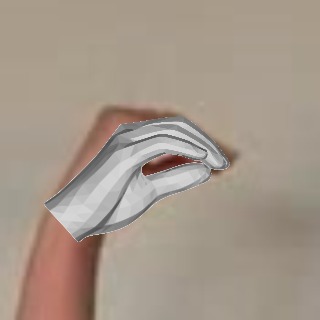}
\end{subfigure}
\begin{subfigure}[c]{\sze\linewidth}
\includegraphics[width=\linewidth]{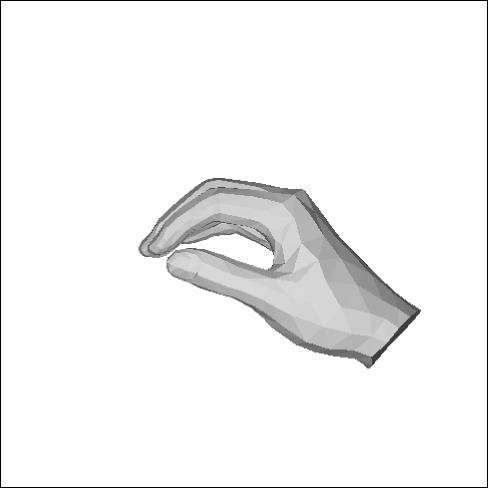}
\end{subfigure}
\begin{subfigure}[c]{\sze\linewidth}
\includegraphics[width=\linewidth]{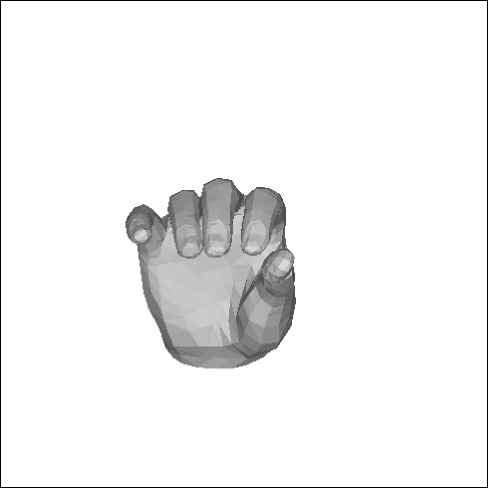}
\end{subfigure}
\begin{subfigure}[c]{\sze\linewidth}
\includegraphics[width=\linewidth]{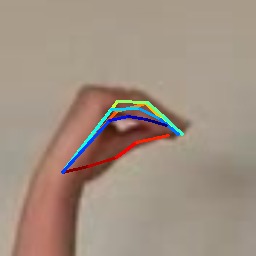}
\end{subfigure}
\hspace{3pt}
\begin{subfigure}[c]{\sze\linewidth}
\includegraphics[width=\linewidth]{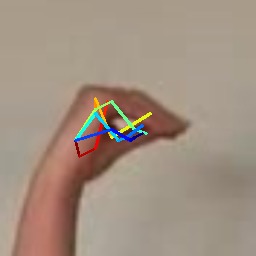}
\end{subfigure}
\begin{subfigure}[c]{\szle\linewidth}
\includegraphics[width=\linewidth]{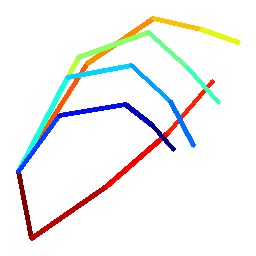}
\end{subfigure}
\hspace{3pt}
\begin{subfigure}[c]{\szle\linewidth}
\includegraphics[width=\linewidth]{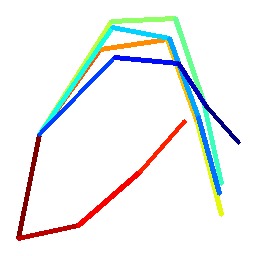}
\end{subfigure}

\begin{subfigure}[c]{\sze\linewidth}
\includegraphics[width=\linewidth]{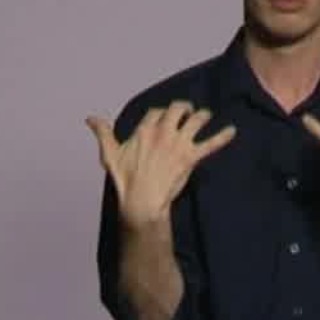}
\end{subfigure}
\begin{subfigure}[c]{\sze\linewidth}
\includegraphics[width=\linewidth]{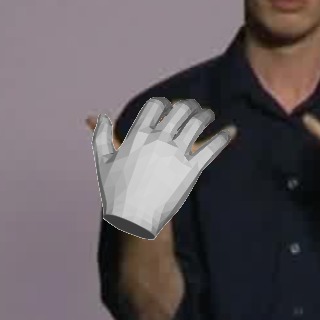}
\end{subfigure}
\begin{subfigure}[c]{\sze\linewidth}
\includegraphics[width=\linewidth]{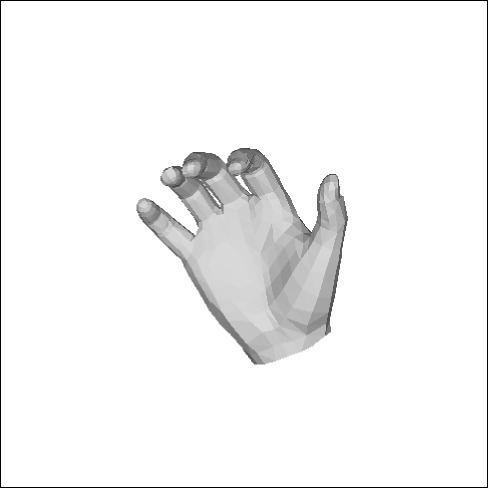}
\end{subfigure}
\begin{subfigure}[c]{\sze\linewidth}
\includegraphics[width=\linewidth]{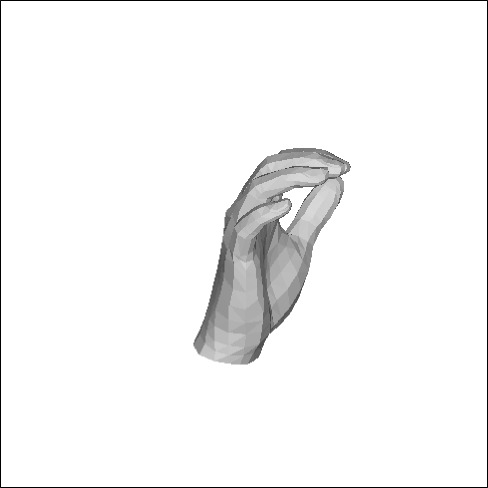}
\end{subfigure}
\begin{subfigure}[c]{\sze\linewidth}
\includegraphics[width=\linewidth]{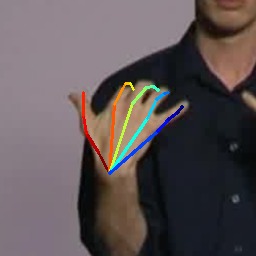}
\end{subfigure}
\hspace{3pt}
\begin{subfigure}[c]{\sze\linewidth}
\includegraphics[width=\linewidth]{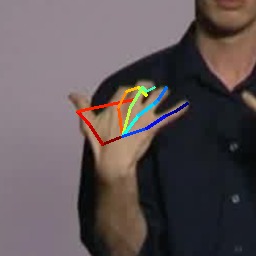}
\end{subfigure}
\begin{subfigure}[c]{\szle\linewidth}
\includegraphics[width=\linewidth]{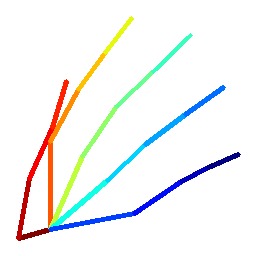}
\end{subfigure}
\hspace{3pt}
\begin{subfigure}[c]{\szle\linewidth}
\includegraphics[width=\linewidth]{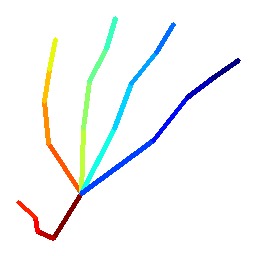}
\end{subfigure}

\begin{subfigure}[c]{\sze\linewidth}
\includegraphics[width=\linewidth]{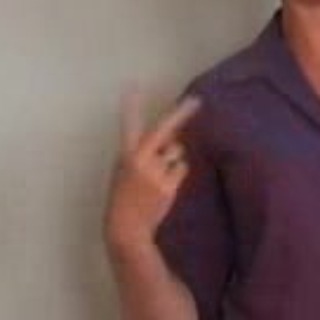}
\end{subfigure}
\begin{subfigure}[c]{\sze\linewidth}
\includegraphics[width=\linewidth]{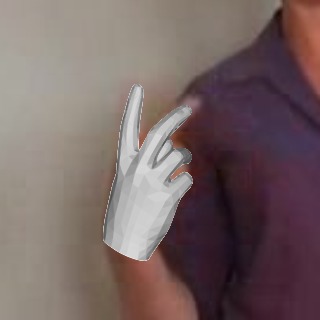}
\end{subfigure}
\begin{subfigure}[c]{\sze\linewidth}
\includegraphics[width=\linewidth]{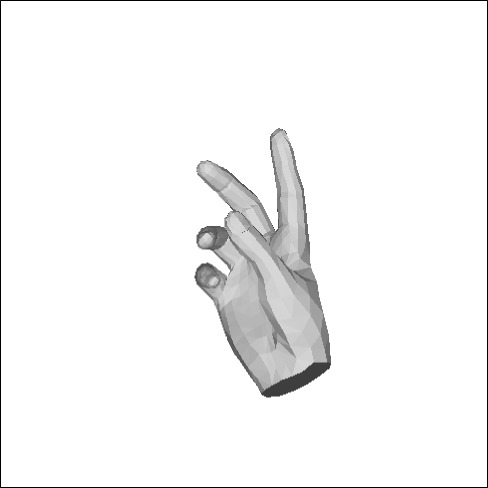}
\end{subfigure}
\begin{subfigure}[c]{\sze\linewidth}
\includegraphics[width=\linewidth]{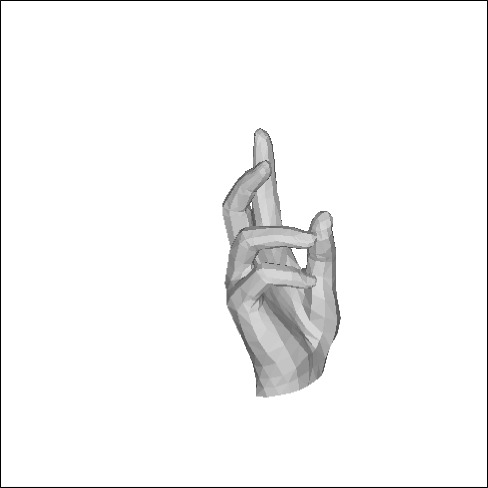}
\end{subfigure}
\begin{subfigure}[c]{\sze\linewidth}
\includegraphics[width=\linewidth]{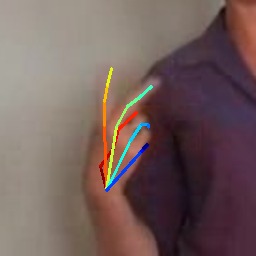}
\end{subfigure}
\hspace{3pt}
\begin{subfigure}[c]{\sze\linewidth}
\includegraphics[width=\linewidth]{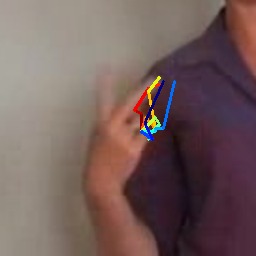}
\end{subfigure}
\begin{subfigure}[c]{\szle\linewidth}
\includegraphics[width=\linewidth]{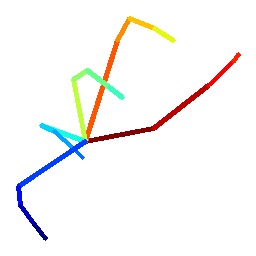}
\end{subfigure}
\hspace{3pt}
\begin{subfigure}[c]{\szle\linewidth}
\includegraphics[width=\linewidth]{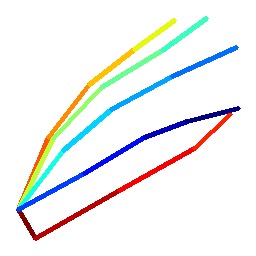}
\end{subfigure}

\begin{subfigure}[c]{\sze\linewidth}
\includegraphics[width=\linewidth]{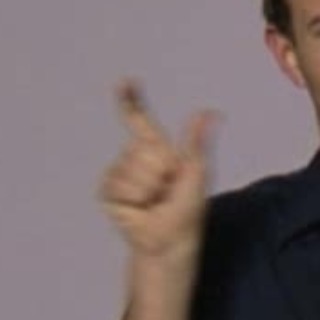}
\end{subfigure}
\begin{subfigure}[c]{\sze\linewidth}
\includegraphics[width=\linewidth]{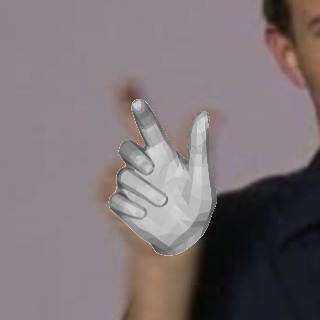}
\end{subfigure}
\begin{subfigure}[c]{\sze\linewidth}
\includegraphics[width=\linewidth]{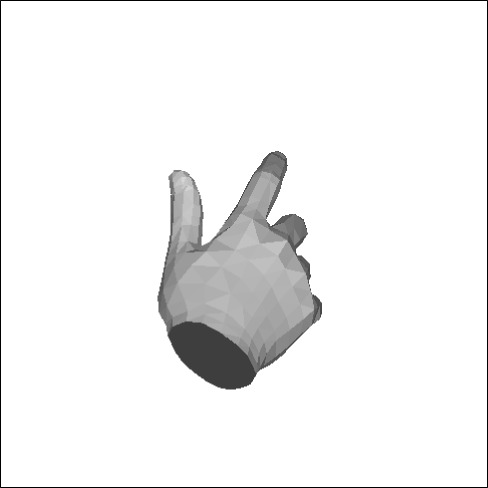}
\end{subfigure}
\begin{subfigure}[c]{\sze\linewidth}
\includegraphics[width=\linewidth]{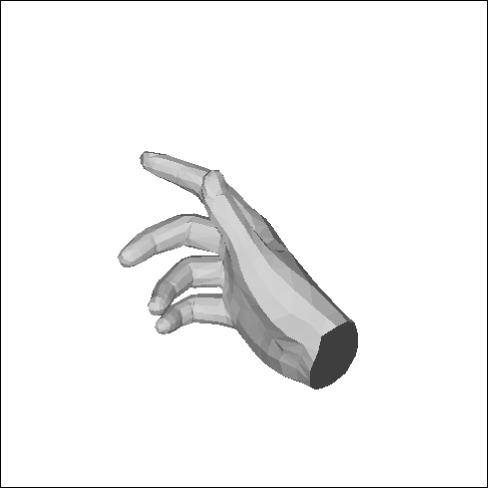}
\end{subfigure}
\begin{subfigure}[c]{\sze\linewidth}
\includegraphics[width=\linewidth]{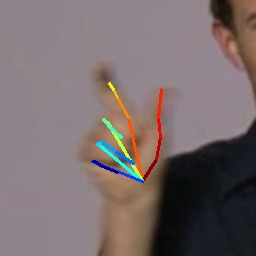}
\end{subfigure}
\hspace{3pt}
\begin{subfigure}[c]{\sze\linewidth}
\includegraphics[width=\linewidth]{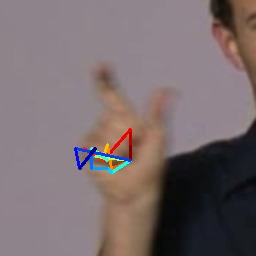}
\end{subfigure}
\begin{subfigure}[c]{\szle\linewidth}
\includegraphics[width=\linewidth]{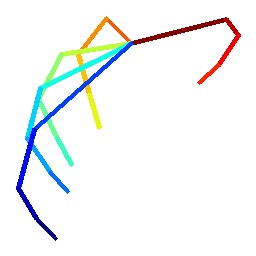}
\end{subfigure}
\hspace{3pt}
\begin{subfigure}[c]{\szle\linewidth}
\includegraphics[width=\linewidth]{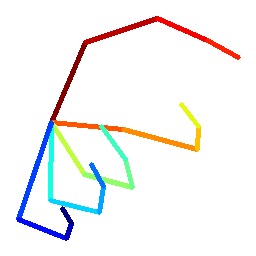}
\end{subfigure}

\begin{subfigure}[c]{\sze\linewidth}
\includegraphics[width=\linewidth]{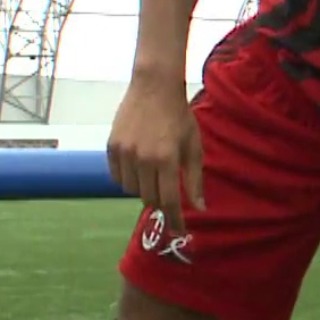}
\caption{Input}
\end{subfigure}
\begin{subfigure}[c]{\sze\linewidth}
\includegraphics[width=\linewidth]{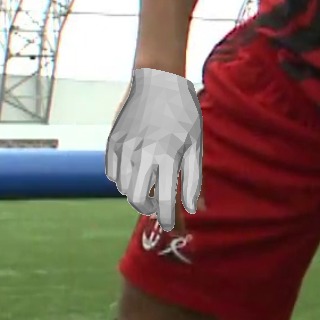}
\caption{Our mesh}
\end{subfigure}
\begin{subfigure}[c]{\sze\linewidth}
\includegraphics[width=\linewidth]{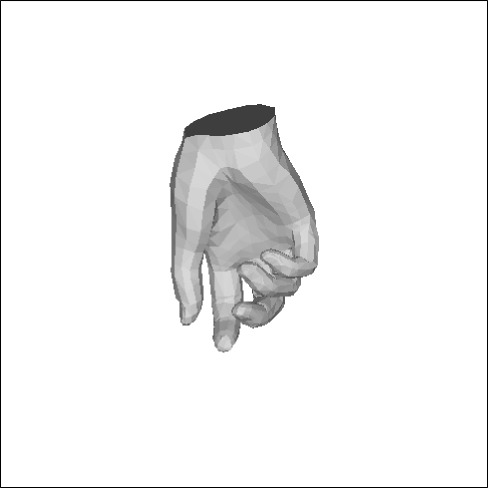}
\caption{Back view}
\end{subfigure}
\begin{subfigure}[c]{\sze\linewidth}
\includegraphics[width=\linewidth]{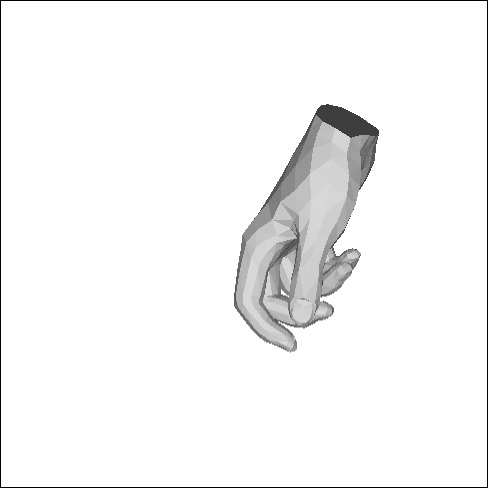}
\caption{Side view}
\end{subfigure}
\begin{subfigure}[c]{\sze\linewidth}
\includegraphics[width=\linewidth]{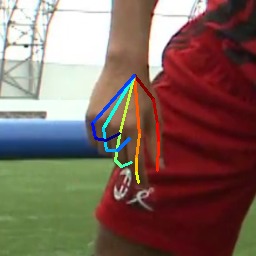}
\caption{Our skeleton}
\end{subfigure}
\hspace{3pt}
\begin{subfigure}[c]{\sze\linewidth}
\includegraphics[width=\linewidth]{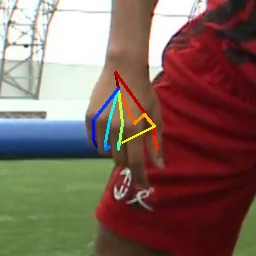}
\caption{\cite{zimmermann2017learning}2D}
\end{subfigure}
\begin{subfigure}[c]{\szle\linewidth}
\includegraphics[width=\linewidth]{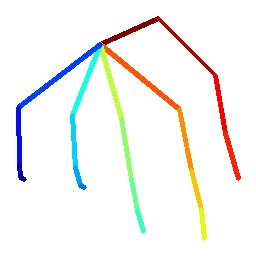}
\caption{\cite{zimmermann2017learning}3D}
\end{subfigure}
\hspace{3pt}
\begin{subfigure}[c]{\szle\linewidth}
\includegraphics[width=\linewidth]{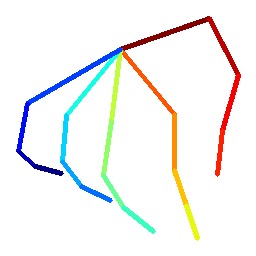}
\caption{\cite{spurr2018cross}3D}
\end{subfigure}

\end{figure*}

\begin{figure*}

\center

\begin{subfigure}[c]{\sze\linewidth}
\includegraphics[width=\linewidth]{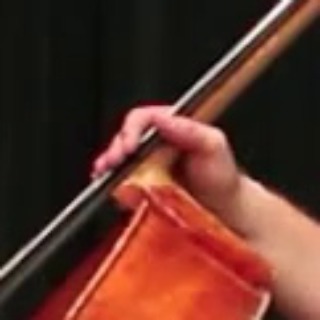}
\end{subfigure}
\begin{subfigure}[c]{\sze\linewidth}
\includegraphics[width=\linewidth]{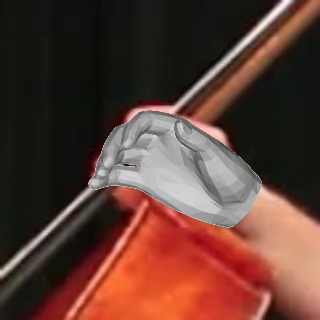}
\end{subfigure}
\begin{subfigure}[c]{\sze\linewidth}
\includegraphics[width=\linewidth]{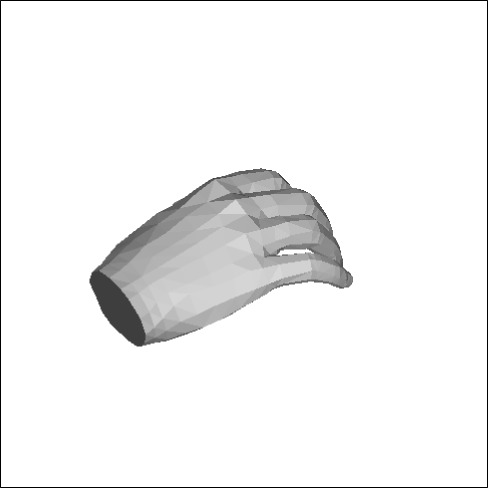}
\end{subfigure}
\begin{subfigure}[c]{\sze\linewidth}
\includegraphics[width=\linewidth]{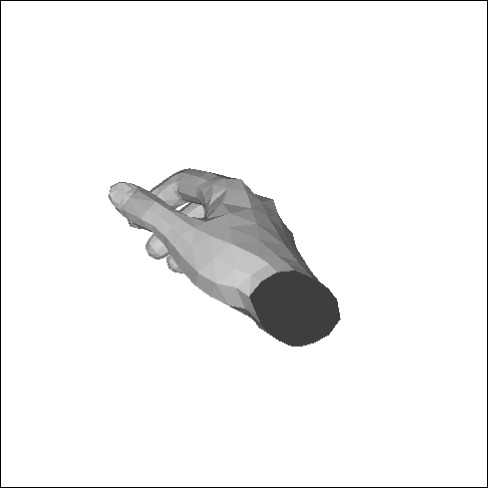}
\end{subfigure}
\begin{subfigure}[c]{\sze\linewidth}
\includegraphics[width=\linewidth]{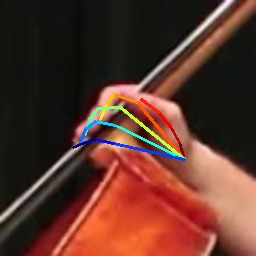}
\end{subfigure}
\hspace{3pt}
\begin{subfigure}[c]{\sze\linewidth}
\includegraphics[width=\linewidth]{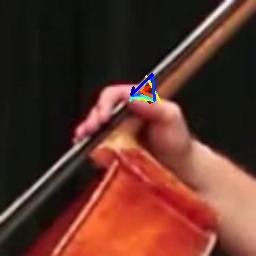}
\end{subfigure}
\begin{subfigure}[c]{\szle\linewidth}
\includegraphics[width=\linewidth]{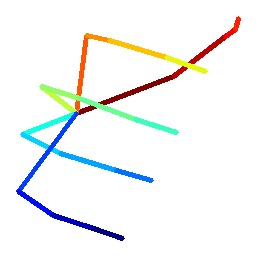}
\end{subfigure}
\hspace{3pt}
\begin{subfigure}[c]{\szle\linewidth}
\includegraphics[width=\linewidth]{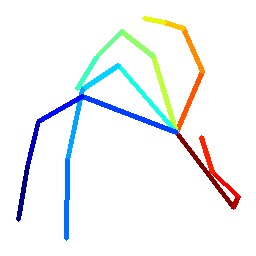}
\end{subfigure}

\begin{subfigure}[c]{\sze\linewidth}
\includegraphics[width=\linewidth]{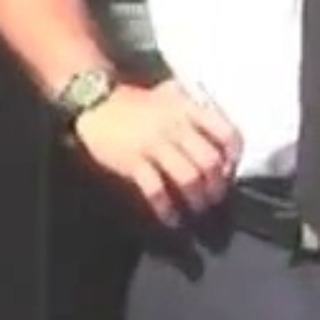}
\end{subfigure}
\begin{subfigure}[c]{\sze\linewidth}
\includegraphics[width=\linewidth]{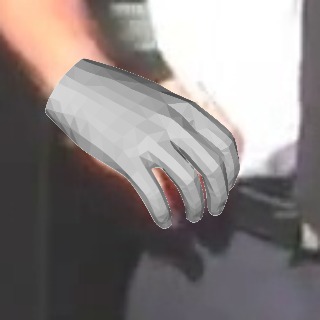}
\end{subfigure}
\begin{subfigure}[c]{\sze\linewidth}
\includegraphics[width=\linewidth]{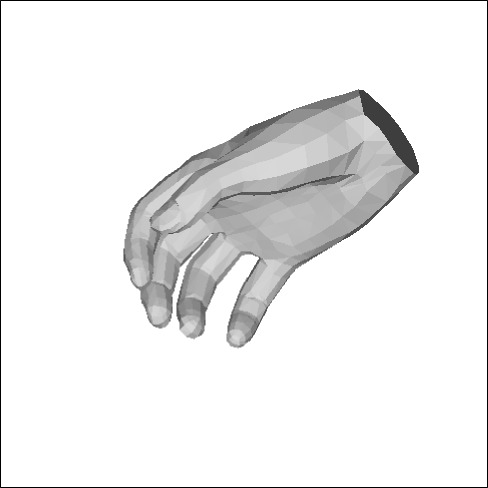}
\end{subfigure}
\begin{subfigure}[c]{\sze\linewidth}
\includegraphics[width=\linewidth]{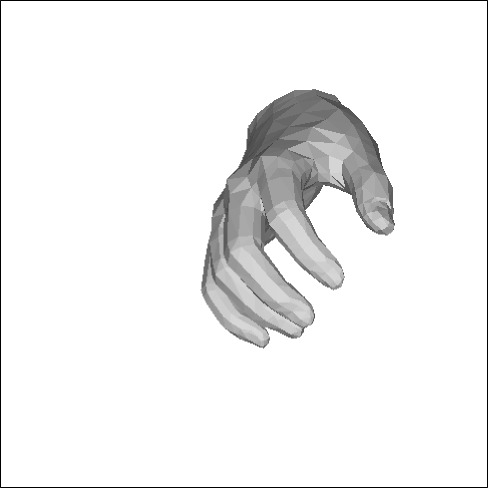}
\end{subfigure}
\begin{subfigure}[c]{\sze\linewidth}
\includegraphics[width=\linewidth]{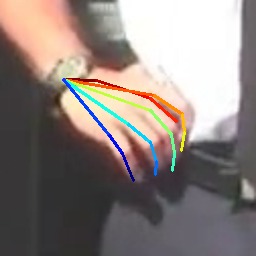}
\end{subfigure}
\hspace{3pt}
\begin{subfigure}[c]{\sze\linewidth}
\includegraphics[width=\linewidth]{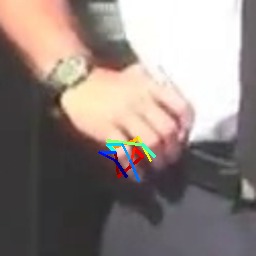}
\end{subfigure}
\begin{subfigure}[c]{\szle\linewidth}
\includegraphics[width=\linewidth]{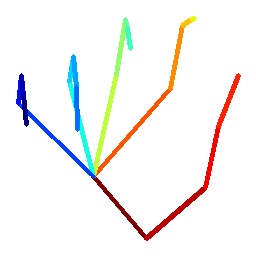}
\end{subfigure}
\hspace{3pt}
\begin{subfigure}[c]{\szle\linewidth}
\includegraphics[width=\linewidth]{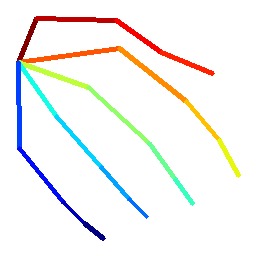}
\end{subfigure}

\begin{subfigure}[c]{\sze\linewidth}
\includegraphics[width=\linewidth]{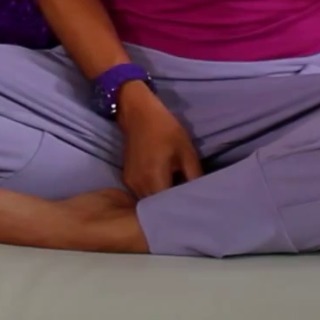}
\end{subfigure}
\begin{subfigure}[c]{\sze\linewidth}
\includegraphics[width=\linewidth]{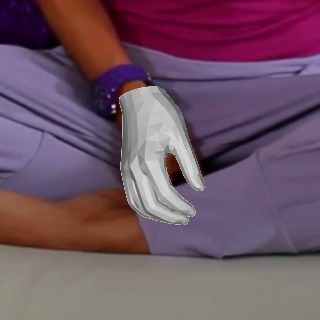}
\end{subfigure}
\begin{subfigure}[c]{\sze\linewidth}
\includegraphics[width=\linewidth]{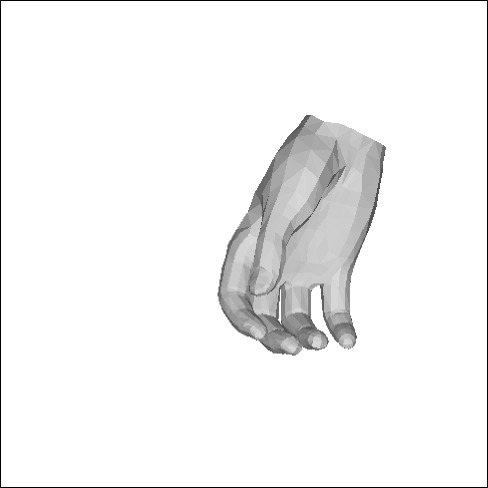}
\end{subfigure}
\begin{subfigure}[c]{\sze\linewidth}
\includegraphics[width=\linewidth]{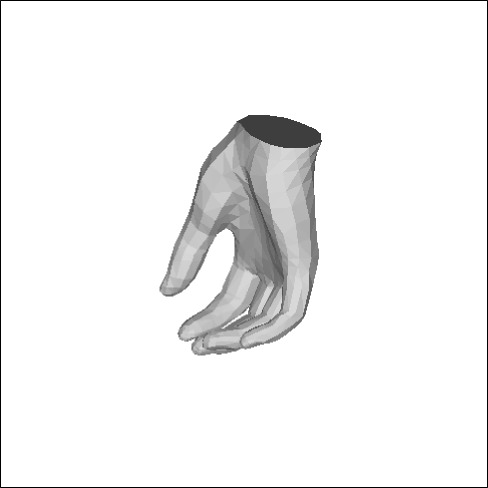}
\end{subfigure}
\begin{subfigure}[c]{\sze\linewidth}
\includegraphics[width=\linewidth]{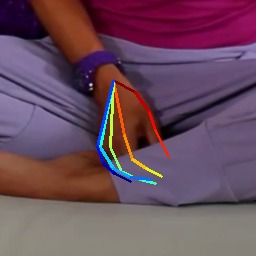}
\end{subfigure}
\hspace{3pt}
\begin{subfigure}[c]{\sze\linewidth}
\includegraphics[width=\linewidth]{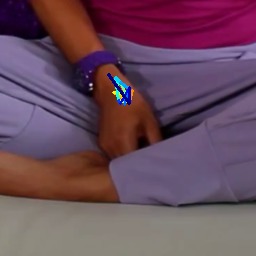}
\end{subfigure}
\begin{subfigure}[c]{\szle\linewidth}
\includegraphics[width=\linewidth]{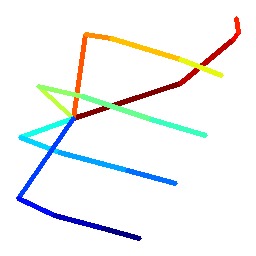}
\end{subfigure}
\hspace{3pt}
\begin{subfigure}[c]{\szle\linewidth}
\includegraphics[width=\linewidth]{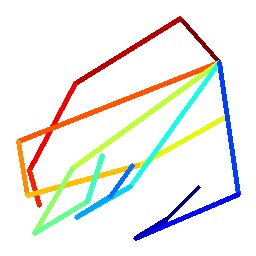}
\end{subfigure}

\begin{subfigure}[c]{\sze\linewidth}
\includegraphics[width=\linewidth]{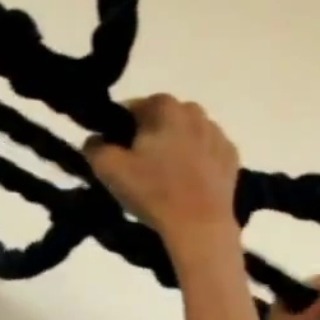}
\end{subfigure}
\begin{subfigure}[c]{\sze\linewidth}
\includegraphics[width=\linewidth]{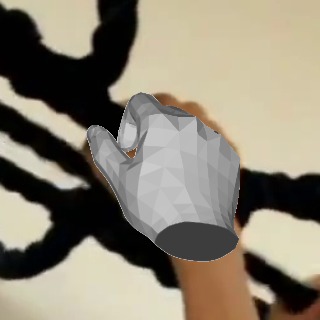}
\end{subfigure}
\begin{subfigure}[c]{\sze\linewidth}
\includegraphics[width=\linewidth]{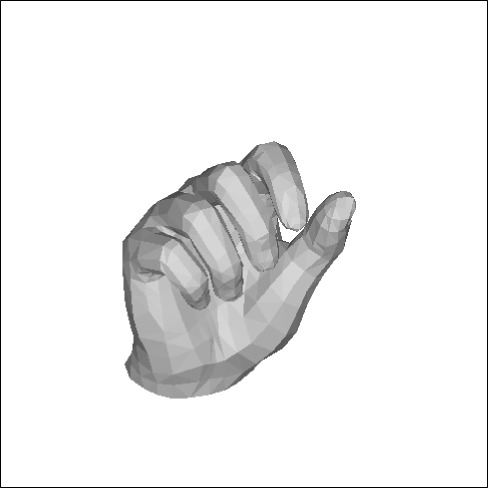}
\end{subfigure}
\begin{subfigure}[c]{\sze\linewidth}
\includegraphics[width=\linewidth]{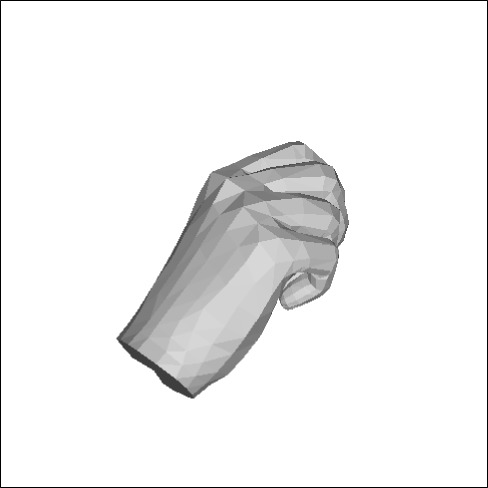}
\end{subfigure}
\begin{subfigure}[c]{\sze\linewidth}
\includegraphics[width=\linewidth]{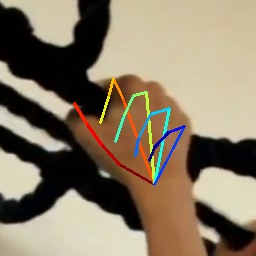}
\end{subfigure}
\hspace{3pt}
\begin{subfigure}[c]{\sze\linewidth}
\includegraphics[width=\linewidth]{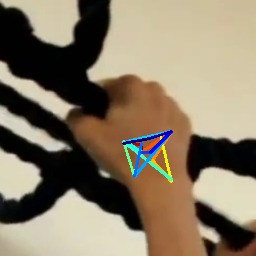}
\end{subfigure}
\begin{subfigure}[c]{\szle\linewidth}
\includegraphics[width=\linewidth]{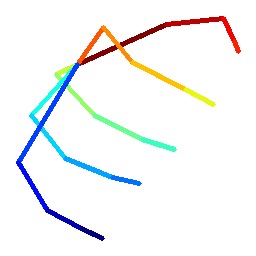}
\end{subfigure}
\hspace{3pt}
\begin{subfigure}[c]{\szle\linewidth}
\includegraphics[width=\linewidth]{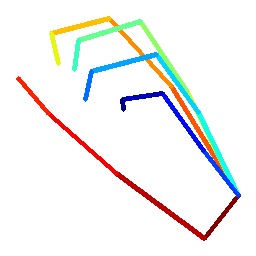}
\end{subfigure}

\begin{subfigure}[c]{\sze\linewidth}
\includegraphics[width=\linewidth]{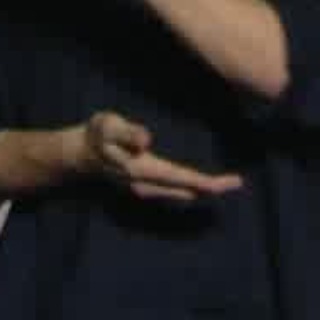}
\end{subfigure}
\begin{subfigure}[c]{\sze\linewidth}
\includegraphics[width=\linewidth]{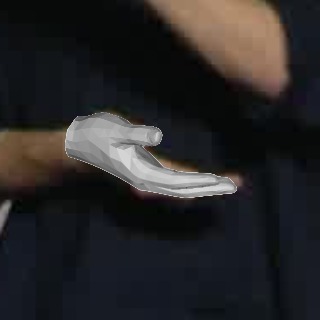}
\end{subfigure}
\begin{subfigure}[c]{\sze\linewidth}
\includegraphics[width=\linewidth]{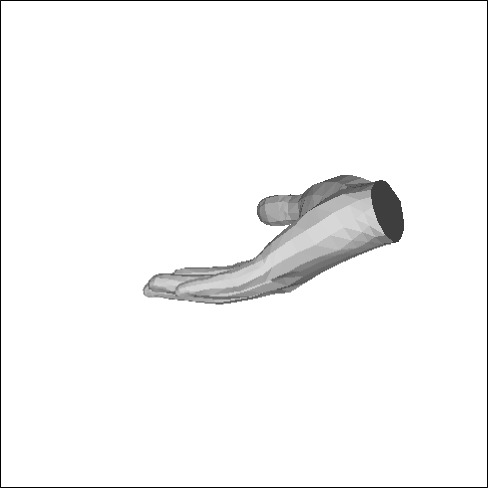}
\end{subfigure}
\begin{subfigure}[c]{\sze\linewidth}
\includegraphics[width=\linewidth]{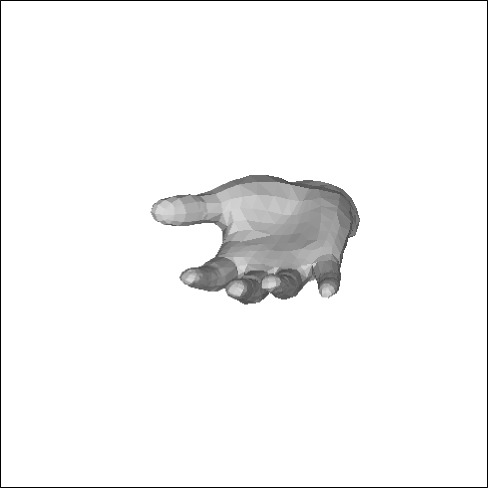}
\end{subfigure}
\begin{subfigure}[c]{\sze\linewidth}
\includegraphics[width=\linewidth]{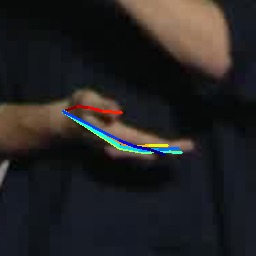}
\end{subfigure}
\hspace{3pt}
\begin{subfigure}[c]{\sze\linewidth}
\includegraphics[width=\linewidth]{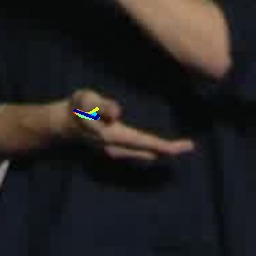}
\end{subfigure}
\begin{subfigure}[c]{\szle\linewidth}
\includegraphics[width=\linewidth]{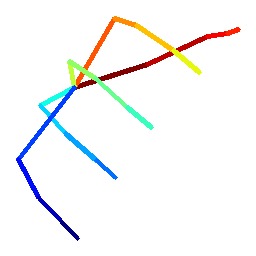}
\end{subfigure}
\hspace{3pt}
\begin{subfigure}[c]{\szle\linewidth}
\includegraphics[width=\linewidth]{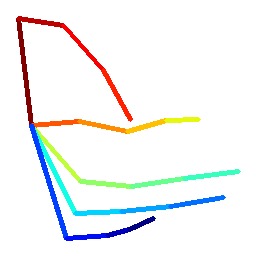}
\end{subfigure}

\begin{subfigure}[c]{\sze\linewidth}
\includegraphics[width=\linewidth]{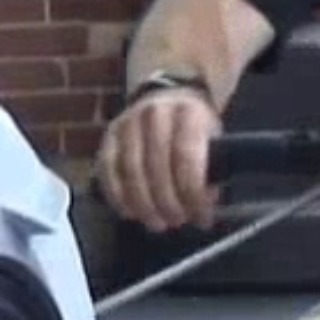}
\end{subfigure}
\begin{subfigure}[c]{\sze\linewidth}
\includegraphics[width=\linewidth]{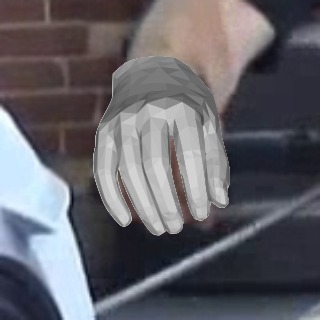}
\end{subfigure}
\begin{subfigure}[c]{\sze\linewidth}
\includegraphics[width=\linewidth]{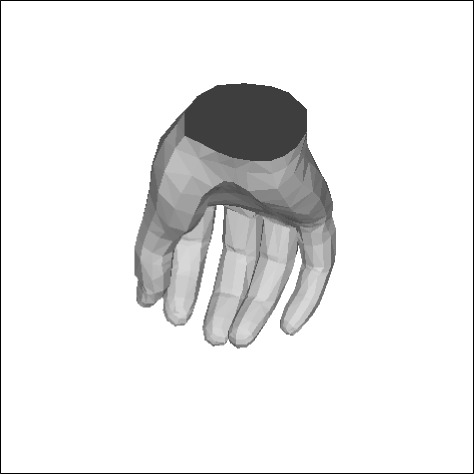}
\end{subfigure}
\begin{subfigure}[c]{\sze\linewidth}
\includegraphics[width=\linewidth]{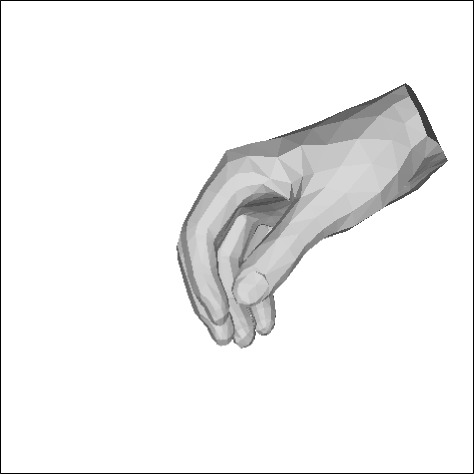}
\end{subfigure}
\begin{subfigure}[c]{\sze\linewidth}
\includegraphics[width=\linewidth]{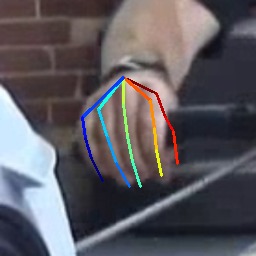}
\end{subfigure}
\hspace{3pt}
\begin{subfigure}[c]{\sze\linewidth}
\includegraphics[width=\linewidth]{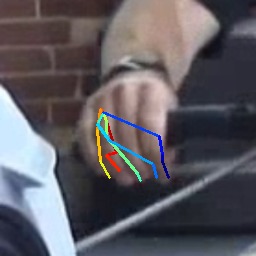}
\end{subfigure}
\begin{subfigure}[c]{\szle\linewidth}
\includegraphics[width=\linewidth]{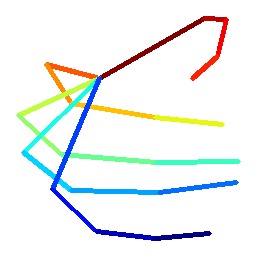}
\end{subfigure}
\hspace{3pt}
\begin{subfigure}[c]{\szle\linewidth}
\includegraphics[width=\linewidth]{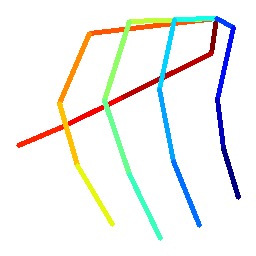}
\end{subfigure}

\begin{subfigure}[c]{\sze\linewidth}
\includegraphics[width=\linewidth]{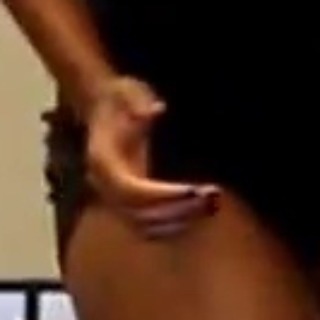}
\end{subfigure}
\begin{subfigure}[c]{\sze\linewidth}
\includegraphics[width=\linewidth]{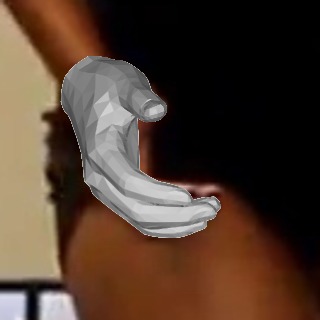}
\end{subfigure}
\begin{subfigure}[c]{\sze\linewidth}
\includegraphics[width=\linewidth]{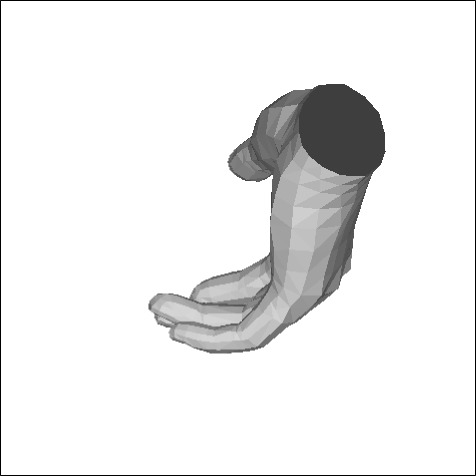}
\end{subfigure}
\begin{subfigure}[c]{\sze\linewidth}
\includegraphics[width=\linewidth]{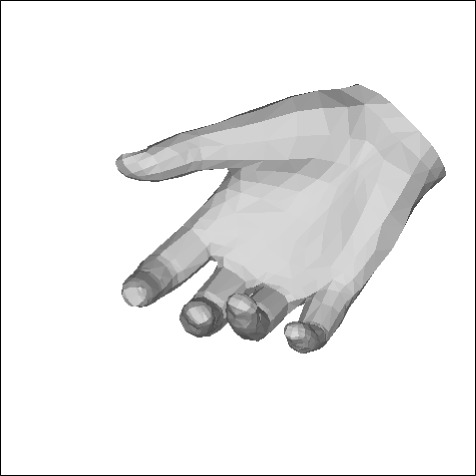}
\end{subfigure}
\begin{subfigure}[c]{\sze\linewidth}
\includegraphics[width=\linewidth]{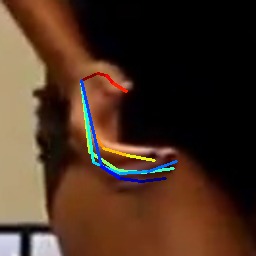}
\end{subfigure}
\hspace{3pt}
\begin{subfigure}[c]{\sze\linewidth}
\includegraphics[width=\linewidth]{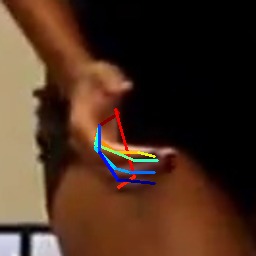}
\end{subfigure}
\begin{subfigure}[c]{\szle\linewidth}
\includegraphics[width=\linewidth]{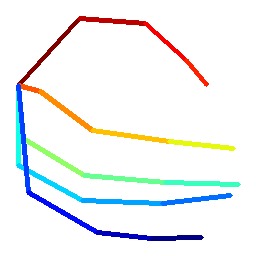}
\end{subfigure}
\hspace{3pt}
\begin{subfigure}[c]{\szle\linewidth}
\includegraphics[width=\linewidth]{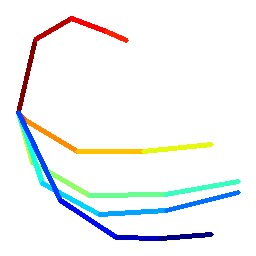}
\end{subfigure}

\begin{subfigure}[c]{\sze\linewidth}
\includegraphics[width=\linewidth]{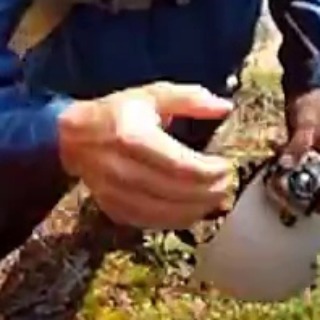}
\caption{Input}
\end{subfigure}
\begin{subfigure}[c]{\sze\linewidth}
\includegraphics[width=\linewidth]{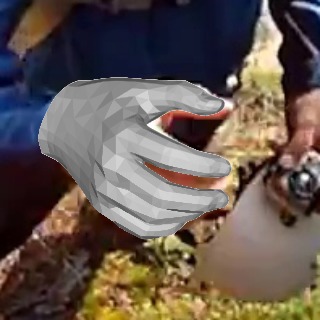}
\caption{Our mesh}
\end{subfigure}
\begin{subfigure}[c]{\sze\linewidth}
\includegraphics[width=\linewidth]{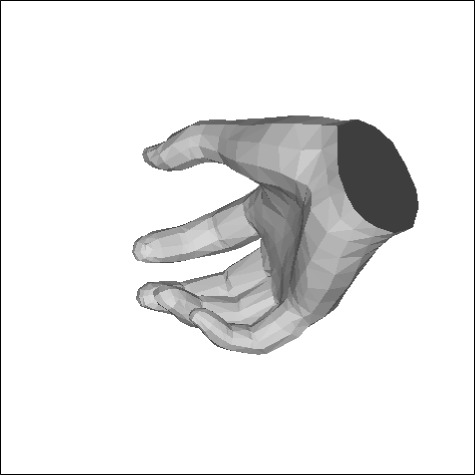}
\caption{Back view}
\end{subfigure}
\begin{subfigure}[c]{\sze\linewidth}
\includegraphics[width=\linewidth]{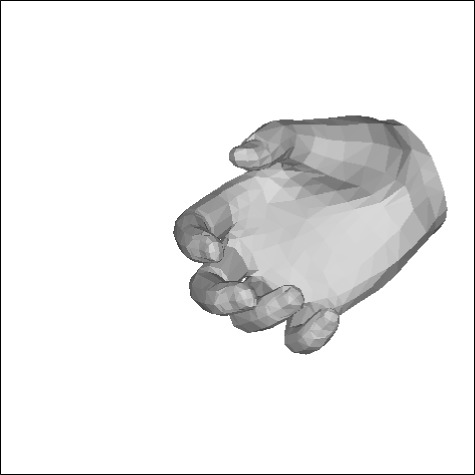}
\caption{Side view}
\end{subfigure}
\begin{subfigure}[c]{\sze\linewidth}
\includegraphics[width=\linewidth]{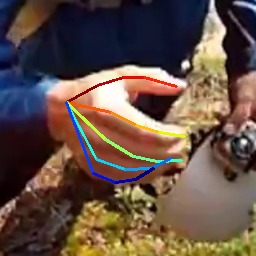}
\caption{Our skeleton}
\end{subfigure}
\hspace{3pt}
\begin{subfigure}[c]{\sze\linewidth}
\includegraphics[width=\linewidth]{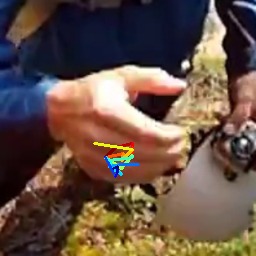}
\caption{\cite{zimmermann2017learning}2D}
\end{subfigure}
\begin{subfigure}[c]{\szle\linewidth}
\includegraphics[width=\linewidth]{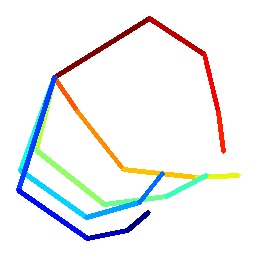}
\caption{\cite{zimmermann2017learning}3D}
\end{subfigure}
\hspace{3pt}
\begin{subfigure}[c]{\szle\linewidth}
\includegraphics[width=\linewidth]{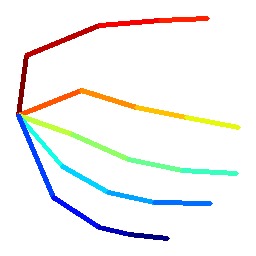}
\caption{\cite{spurr2018cross}3D}
\end{subfigure}

\caption*{Our 3D hand reconstruction on examples from the challenging testing set of \textsc{Mpii+Nzsl} compared to the 3D hand pose predictions of \cite{zimmermann2017learning} and \cite{spurr2018cross}.}

\end{figure*}

\end{document}